\documentclass[10pt,twocolumn,letterpaper]{article}

\usepackage[pagenumbers]{cvpr} % To force page numbers, e.g. for an arXiv version

\usepackage{graphicx}
\usepackage{amsmath}
\usepackage{amssymb}
\usepackage{booktabs}
\usepackage{float}
\usepackage{placeins}
\usepackage[table,x11names]{xcolor}
\definecolor{maroon}{cmyk}{0,0.87,0.68,0.32}

\makeatletter
\newcommand\pagenumberingnoreset[1]{\gdef\thepage{\csname @#1\endcsname\c@page}}
\makeatother

\usepackage{xcolor,amsmath}
\usepackage[linesnumbered,ruled,vlined]{algorithm2e}
\DontPrintSemicolon

\SetKwComment{Comment}{\color{green!50!black}// }{}

\SetKwProg{Function}{function}{}{}

\usepackage[pagebackref,breaklinks,colorlinks]{hyperref}
\usepackage[accsupp]{axessibility}

\usepackage[capitalize]{cleveref}
\crefname{section}{Sec.}{Secs.}
\Crefname{section}{Section}{Sections}
\Crefname{table}{Table}{Tables}
\crefname{table}{Tab.}{Tabs.}

\newcommand\blfootnote[1]{%
  \begingroup
  \renewcommand\thefootnote{}\footnote{#1}%
  \addtocounter{footnote}{-1}%
  \endgroup
}

\def\cvprPaperID{11604} % *** Enter the CVPR Paper ID here
\def\confName{CVPR}
\def\confYear{2022}
\usepackage{overpic}
\usepackage{enumitem} %< control spacing in itemize/enumerate/...
\usepackage{overpic} %< add raw math symbols to figures
\usepackage{color}
\definecolor{turquoise}{cmyk}{0.65,0,0.1,0.3}
\definecolor{purple}{rgb}{0.65,0,0.65}
\definecolor{dark_green}{rgb}{0, 0.5, 0}
\definecolor{orange}{rgb}{0.8, 0.6, 0.2}
\definecolor{red}{rgb}{0.8, 0.2, 0.2}
\definecolor{darkred}{rgb}{0.6, 0.1, 0.05}
\definecolor{blueish}{rgb}{0.0, 0.3, .6}
\definecolor{light_gray}{rgb}{0.7, 0.7, .7}
\definecolor{pink}{rgb}{1, 0, 1}
\definecolor{greyblue}{rgb}{0.25, 0.25, 1}

\usepackage{blindtext}

\renewcommand{\paragraph}[1]{\vspace{1em}\noindent\textbf{#1}.}
\begin{document}
%\iffalse

\title{Scaling Vision Transformers to Gigapixel Images via \\ Hierarchical Self-Supervised Learning}

\author{Richard J. Chen$^{1}$, Chengkuan Chen$^{1}$, Yicong Li$^{1}$, Tiffany Y. Chen$^{1}$, \\ Andrew D. Trister$^{2}$, Rahul G. Krishnan$^{3,*}$, Faisal Mahmood$^{1,*}$\\
$^{1}$Harvard, BWH, Broad Institute \enspace $^{2}$Bill \& Melinda Gates Foundation \enspace $^{3}$University of Toronto\\
{\tt\small {richardjchen@g.harvard.edu, faisalmahmood@bwh.harvard.edu}}
}
\maketitle
\begin{abstract}
Vision Transformers (ViTs) and their multi-scale and hierarchical variations have been successful at capturing image representations but their use has been generally studied for low-resolution images (\textit{e.g.} $256 \times 256, 384 \times 384$). For gigapixel whole-slide imaging (WSI) in computational pathology, WSIs can be as large as $150000 \times 150000$ pixels at $20\times$ magnification and exhibit a hierarchical structure of visual tokens across varying resolutions: from $16 \times 16$ images capturing individual cells, to $4096 \times 4096$ images characterizing interactions within the tissue microenvironment. We introduce a new ViT architecture called the Hierarchical Image Pyramid Transformer (HIPT), which leverages the natural hierarchical structure inherent in WSIs using two levels of self-supervised learning to learn high-resolution image representations. HIPT is pretrained across 33 cancer types using 10,678 gigapixel WSIs, 408,218 $4096 \times 4096$ images, and 104M $256 \times 256$ images. We benchmark HIPT representations on 9 slide-level tasks, and demonstrate that: 1) HIPT with hierarchical pretraining outperforms current state-of-the-art methods for cancer subtyping and survival prediction, 2) self-supervised ViTs are able to model important inductive biases about the hierarchical structure of phenotypes in the tumor microenvironment.

%To model this hierarchy, we interpret WSIs essentially as a "long document" in natural language processing (NLP). Similar to how hierarchical networks in NLP aggregate word-level embeddings to form sentence-level embeddings, followed by sentence-level aggregation to form the docoument-level embedding, we develop a hierarchical architecture that aggregates $16 \times 16$-level visual tokens in their respective $256 \times 256$ and $4096 \times 4096$ windows to eventually form the slide-level representation.

% we perform $4096$-level patching of the WSI as a sequence of "regions", with subsequent $256$-level patching of regions as sequences of "patches" and then $16$-level patching of patches as sequences of "subpatches".

% In computational pathology, conventional approaches for slide-level supervision use Multiple Instance Learning (MIL) in patching at a fixed resolution of $256 \times 256$, followed by only global pooling due to the large sequence lengths in an unrolled WSI. To address these challenges

\end{abstract}
\vspace{-4mm}
\section{Introduction}
\label{sec:intro}
\begin{figure}[t]
\begin{center}
% \begin{overpic} 
% [width=\linewidth]
% {example-image-a}
% \end{overpic}
\includegraphics[width=\linewidth]{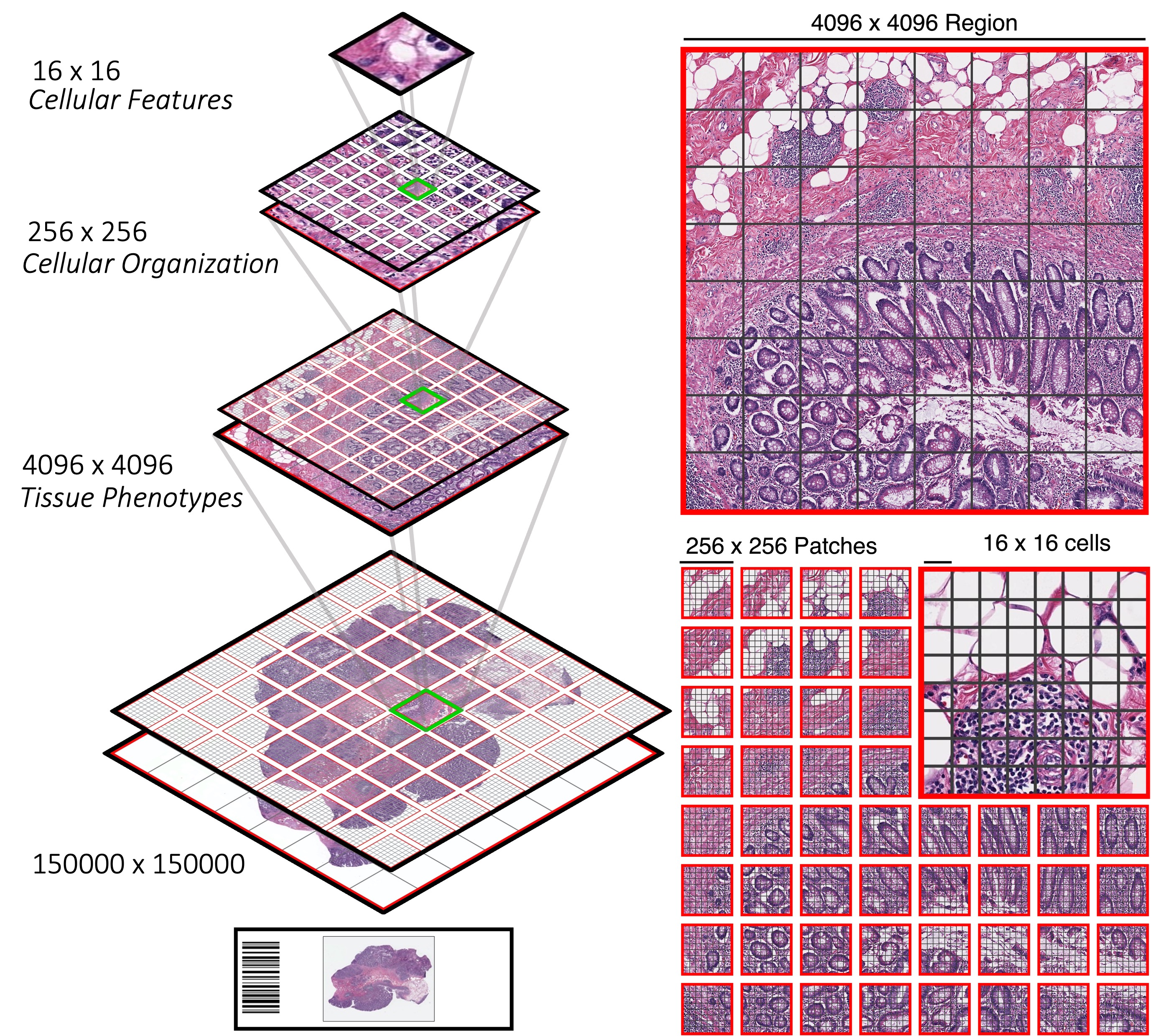}
\end{center}
\caption{
\textbf{Hierarchical Structure of Whole-Slide Images (WSIs).} \textbf{Left.} Unlike natural images, since WSIs have a fixed scale, there exists a hierarchical structure of visual tokens at varying image resolutions. \textbf{Right.} In addition to formulating a single $256 \times 256$ image as as sequence of 256 $[16 \times 16]$ tokens, we can also view these $256 \times 256$ image as being part of a larger, disjoint sequence of $[256 \times 256]$ tokens in a $4096 \times 4096$ region.}
\label{fig:hier_wsi}

\end{figure}
Tissue phenotyping is a fundamental problem in computational pathology (CPATH) that aims at characterizing objective, histopathologic features within gigapixel whole-slide images (WSIs) for cancer diagnosis, prognosis, and the estimation of response-to-treatment in patients~\cite{ludwig2005biomarkers, kather2016multi, javed2020cellular}. Unlike natural images, whole-slide imaging is a challenging computer vision domain in which image resolutions can be as large as $150000 \times 150000$ pixels\blfootnote{$^{*}$ Contributed Equally.}, with many methods using the following three-stage, weakly-supervised framework based on multiple instance learning (MIL): 1) tissue patching at a single magnification objective (``zoom"), 2) patch-level feature extraction to construct a sequence of embedding instances, and 3) global pooling of instances to construct a slide-level representation for weak-supervision using slide-level labels (\textit{e.g.} - subtype, grade, stage, survival, origin)~\cite{hou2016patch, courtiol2019deep, van2019no, tellez2019neural, ilse2018attention, campanella2019clinical, lu2020data, zhu2017wsisa, lu2021ai}.

Though achieving ``clinical-grade" performance on many cancer subtyping and grading tasks, this three-stage process has a few important design limitations. First, patching and feature extraction are generally fixed to $[256 \times 256]$ context regions. Though able to discern fine-grained morphological features such as nuclear atypia or tumor presence, depending on the cancer type, $[256 \times 256]$ windows have limited context in capturing coarser-grained features such as tumor invasion, tumor size, lymphocytic infiltrates, and the broader spatial organization of these phenotypes in the tissue microenvironment, as depicted in Figure \ref{fig:hier_wsi}~\cite{beck2011systematic, Diao2021, chan2019histosegnet}. Second, in contrast with other image-based sequence modeling approaches such as Vision Transformers (ViTs), MIL uses only global pooling operators due to the large sequence lengths of WSIs~\cite{ilse2018attention}. As a result, this limitation precludes the application of Transformer attention for learning long-range dependencies between phenotypes such as tumor-immune localization, an important prognostic feature in survival prediction~\cite{saltz2018spatial, le2020utilizing, abduljabbar2020geospatial}. Lastly, though recent MIL approaches have adopted self-supervised learning as a strategy for patch-level feature extraction (called tokenization in ViT literature), parameters in the aggregation layers still require training~\cite{ciga2021self, saillard2021self, dehaene2020self, koohbanani2021self, boyd2021self, li2021dual, chen2022self}. In viewing patch-based sequence modeling of WSIs in relation to ViTs, we note that the architectural design choice of using Transformer attention enables pretraining of both the tokenization and aggregation layers in ViT models, which is important in preventing MIL models from over- or under-fitting in low-data regimes~\cite{dosovitskiy2021image, li2022efficient, caron2021emerging, bao2022beit, he2021masked}.

To address these issues, we explore the challenge of developing a Vision Transformer for slide-level representation learning in WSIs. In comparison to natural images which are actively explored by ViTs, we note {\textit{a key difference in modeling WSIs is that visual tokens would always be at a fixed scale for a given magnification objective}. For instance, scanning WSIs at a $20\times$ objective results in a fixed scale of approximately $0.5 \mu m$ per pixel, allowing for consistent comparison of visual elements that may elucidate important histomorphological features beyond their normal reference ranges. Moreover, WSIs also exhibit a hierarchical structure of visual tokens at varying image resolutions at $20\times$ magnification: the $16 \times 16$ images encompass the bounding box of cells and other fine-grained features (stroma, tumor cells, lymphocytes)~\cite{Diao2021, hou2019sparse}, $256 \times 256$ images capture local clusters of cell-to-cell interactions (tumor cellularity)~\cite{graham2019hover, bejnordi2017diagnostic, pati2020hact, abousamra2021multi}, $1024 \times 1024$-$4096 \times 4096$ images further characterize macro-scale interactions between clusters of cells and their organization in tissue (the extent of tumor-immune localization in describing tumor-infiltrating versus tumor-distal lymphocytes)~\cite{abduljabbar2020geospatial, brancati2021bracs}, and finally the overall intra-tumoral heterogeneity of the tissue microenvironment depicted at the slide-level of the WSI~\cite{javed2020cellular,saltz2018spatial, hosseini2019atlas, balkwill2012tumor, marusyk2012intra}. The hypothesis that this work tests is that the judicious use of this hierarchy in self-supervised learning results in better slide-level representations.

We introduce a Transformer-based architecture for hierarchical aggregation of visual tokens and pretraining in gigapixel pathology images, called Hierarchical Image Pyramid Transformer (HIPT). We approach the task of slide-level representation learning in a manner similar to learning long document representations in language modeling, in which we develop a three-stage hierarchical architecture that performs bottom-up aggregation from [$16 \times 16$] visual tokens in their respective $256 \times 256$ and $4096 \times 4096$ windows to eventually form the slide-level representation, as demonstrated in Figure \ref{fig:overview}~\cite{yang2016hierarchical, zhang-etal-2019-hibert}. Our work pushes the boundaries of both Vision Transformers and self-supervised learning in two important ways. By modeling WSIs as a disjoint set of nested sequences, within HIPT: 1) we decompose the problem of learning a good representation of a WSI into hierarchically-related representations each of which can be learned via self-supervised learning, and 2) we use student-teacher knowledge distillation (DINO ~\cite{caron2021emerging}) to pretrain each aggregation layers with self-supervised learning on regions as large as $4096 \times 4096$.

% we decompose the problem of 

%In modeling WSIs as a disjoint set of nested, sequences, we 
%\begin{enumerate}
%\vspace{-3mm}
%\item With shorter sequences, we can tractably extend self-attention computation in modeling dependencies between tokens at each layer of the hierarchy. For $[16 \times 16]$ cell aggregation, $[256 \times 256]$ patch aggregation and $4096 \times 4096$ region aggregation, we use three vanilla ViT backbones (denoted as $\operatorname{ViT-16}$, $\operatorname{ViT-256}$, $\operatorname{ViT-4096}$ respectively).
%\vspace{-3mm}
%\item In using Transformer attention on fixed image windows, we can also adopt self-supervised ViT techniques such as  student-teacher knowledge distillation for pretraining the aggregation layers across the different stages of HIPT, which we call hierarchical pretraining.
%\vspace{-3mm}
%\end{enumerate}
We apply HIPT to the task of learning representations of gigapixel histopathological images extracted at $20\times$ resolution. We show that our method achieves superior performance to conventional MIL approaches. The difference is pronounced in context-aware tasks such as survival prediction in which larger context is appreciated in characterizing broader prognostic features in the tissue microenvironment~\cite{pati2022hierarchical, saltz2018spatial, abduljabbar2020geospatial, chen2021multimodal}. Using K-Nearest Neighbors on the $4096 \times 4096$ representations of our model, we outperform several weakly-supervised architectures in slide-level classification -- an important step forward in achieving self-supervised slide-level representations. Finally, akin to self-supervised ViTs on natural images that can perform semantic segmentation of the scene layout, we find that the multi-head self-attention in self-supervised ViTs learn visual concepts in histopathology tissue (from fine-grained visual concepts such as cell locations in the $\operatorname{ViT_{256}\textrm{-}16}$ to coarse-grained visual concepts such as broader tumor cellularity in the $\operatorname{ViT_{4096}\textrm{-}256}$), as demonstrated in Figure \ref{fig:attention_column}, \ref{fig:hierarchical_attention}. We make code available at \href{https://github.com/mahmoodlab/HIPT}{https://github.com/mahmoodlab/HIPT}.

\section{Related Work}
\label{sec:related}

\vspace{-3mm}
\paragraph{Multiple Instance Learning in WSIs} In general set-based deep learning, Edwards \& Storkey and Zaheer \etal proposed the first network architectures operating on set-based data structures, with Brendel \etal demonstrating ``bag-of-features" able to reach high accuracy on ImageNet~\cite{DBLP:conf/iclr/EdwardsS17, zaheer2017deep, DBLP:conf/iclr/BrendelB19}. Concurrently in pathology, Ilse \textit{et al.} extended set-based network architectures as an approach for multiple instance learning in histology region-of-interests, with Campanella \textit{et al.} later extending end-to-end weak-supervision on gigapixel WSIs~\cite{ilse2018attention, campanella2019clinical}. Lu \textit{et al.} demonstrated that by using a pretrained ResNet-50 encoder on ImageNet for instance-level feature extraction, only a global pooling operator needs to be trained for weakly-supervised slide-level tasks~\cite{lu2020data}. Following Lu \textit{et al.}, there have been many variations of MIL that have adapted image pretraining techniques such as VAE-GANs, SimCLR, and MOCO as instance-level feature extraction~\cite{zhao2020predicting, li2021dual, saillard2021self}. Recent variations of MIL have also evolved to extend the aggregation layers and scoring functions~\cite{zhu2017wsisa, xu2019camel, tellez2019neural, shao2021transmil, yao2019deep, yao2020whole, chen2021multimodal}. Li \textit{et al.} proposed a multi-scale MIL approach that performs patching and self-supervised instance learning at $20\times$ and $5\times$ resolution, followed by spatially-resolved alignment of patches\cite{li2021dual}. The integration of magnification objectives within WSIs has been followed in other works as well~\cite{hashimoto2020multi, marini2021multi, meng2021hierarchical, gao2016multi}, however, we note that combining visual tokens across objectives would not share the same scale. In this work, patching is done at a single magnification objective, with larger patch sizes used to capture macro-scale morphological features, which we hope will contribute towards a shift in rethinking context modeling of WSIs.

\paragraph{Vision Transformers and Image Pyramids} The seminal work of Vaswani \etal has led to remarkable developments in not only language modeling, but also image representation learning via Vision Transformers (ViTs), in which $256 \times 256$ images are formulated as an image patch sequence of $[16 \times 16]$ visual tokens~\cite{vaswani2017attention, dosovitskiy2021image, touvron2021training}. Motivated by multiscale, pyramid-based image processing \cite{rosenfeld1971edge, burt1987laplacian, koenderink1984structure}, recent progress in ViT architecture development has focused on efficiency and integration of multiscale information (\textit{e.g.} - Swin, ViL, TNT, PVT, MViT) in addressing the varying scale / aspect ratios of visual tokens~\cite{han2021transformer, zhang2021multi, liu2021swin, wang2021pyramid, fan2021multiscale}. In contrast with pathology, we highlight that learning scale invariance may not be necessary if the image scale is fixed at a given magnification. Similar to our work is NesT and Hierarchical Perciever, which similarly partitions and then aggregates features from non-overlapping image regions via Transformer blocks\cite{zhang2022nested, carreira2022hierarchical}. A key difference is that we show ViT blocks at each stage can be separately pretrained for high-resolution encoding (up to $4096 \times 4096)$.
\begin{figure*}
\begin{center}
\includegraphics[width=1.00\linewidth]{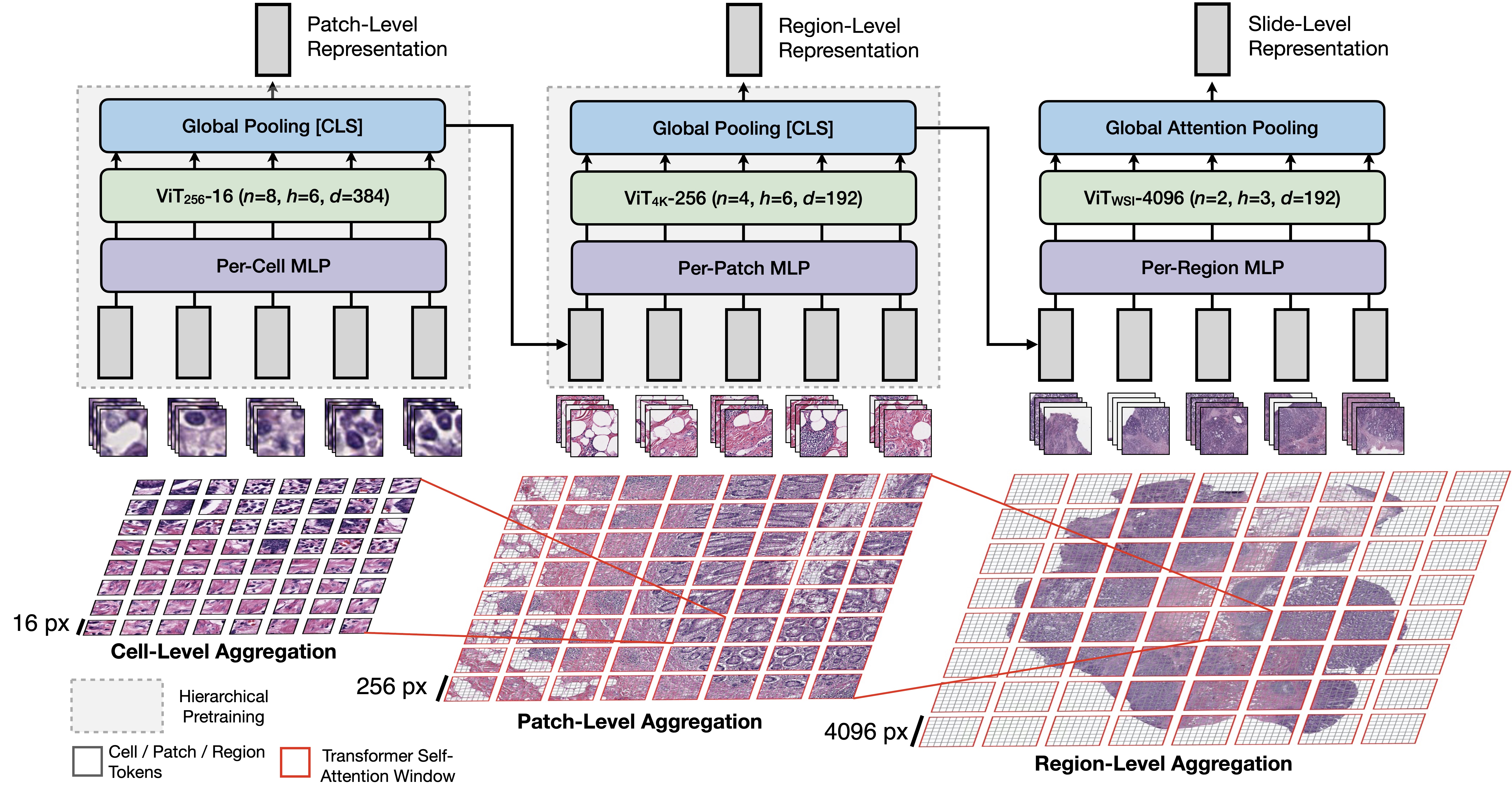}
\end{center}
\vspace{-4mm}
\caption{\small
\textbf{HIPT Architecture.} Motivated by the use of hierarchical representations in natural language processing, where embeddings can be aggregated at the character-, word-, sentence- and paragraph-level to form document representations, we aggregate visual tokens at the $\textbf{x}_{16}$ cell-, $\textbf{x}_{256}$ patch-, $\textbf{x}_{4096}$ region-level to form slide representations. To also model important dependencies between visual concepts at each stage, we adapt Transformer self-attention as a permutation-equivariant aggregation layer. Note that since the complexity of patching $\textbf{x}_{4096}$ regions with $\textbf{x}_{256}$ tokens is the same as patching  $\textbf{x}_{256}$ images with $\textbf{x}_{16}$ tokens, we can pretrain aggregation layers for high-resolution images using similar self-supervised ViT techniques for low-resolution images.}
\label{fig:overview}
\vspace{-4mm}
\end{figure*} %< bloody latex and its heuristics for figure placement

\section{Method}
\subsection{Problem Formulation}
\noindent \textbf{Patch Size and Visual Token Notation:}  We use the following notation to distinguish between the sizes of ``images" and ``tokens" that correspond to that image. For an image $\textbf{x}$ with resolution $L \times L$ (or $\textbf{x}_L$), we refer to sequence of extracted visual tokens from non-overlapping patches (of size $[l \times l]$) within $\textbf{x}_L$ as $\{\textbf{x}_{l}^{(i)}\}_{i=1}^{M} \in \mathbb{R}^{M \times d_{l}}$, where $M$ is the sequence length and $d$ is the embedding dimension extracted for $l$-sized tokens. In working with multiple image resolutions (and their respective tokens) in a WSI, we additionally denote the shape of visual tokens (and the patching parameter) within $\textbf{x}_L$ image as $[l \times l]$ (using brackets). For natural images with size $\textbf{x}_{256}$, ViTs generally use $l = L^{1/2} =16$ which results in a sequence length of $M=256$. Additionally, we denote a ViT working on a $L$-sized image resolution with $[l \times l]$ tokens as $\operatorname{ViT}_L-l$. For $\textbf{x}_{\text{WSI}}$ (referring to the slide-level resolution of the WSI), MIL approaches choose $l = 256$ which fits the input shape of CNN encoders that can be pretrained and using for tokenization, resulting in $M > 10,000$ (variable due to the total area of segmented tissue content).

\noindent \textbf{Slide-Level Weak Supervision:} For a WSI $\textbf{x}_{\text{WSI}}$ with outcome $y$, the goal is to solve the slide-level classification task $P(y|\textbf{x}_{\text{WSI}})$. Conventional approaches for solving this task use a three-stage MIL framework which performs: 1) $[256 \times 256]$-patching, 2) tokenization, and 3) global attention pooling.  $\textbf{x}_{\text{WSI}}$ is formulated as the sequence $\{\textbf{x}_{256}^{(i)}\}_{i=1}^{M} \in \mathbb{R}^{M \times 1024}$ which results from using a ResNet-50 encoder pretrained on ImageNet (truncated after the 3rd residual block). Due to the large sequence lengths with $l=256$, neural network architectures in this task are limited to per-patch and global pooling operators in extracting a slide-level embedding for downstream tasks. % A more thorough overview on MIL is given in the \textbf{Supplement}.

\iffalse
with the general form $\mathcal{F}$:
\begin{equation}
\mathcal{F}\left(X\right)= \zeta \big(\rho ( \{ \phi ( \mathbf{x}^{(i)}_{256} ): \mathbf{x}^{(i)}_{256} \in \textbf{x}_{\text{WSI}} \} ) \big)
\end{equation}
in which $\phi: \mathbb{R}^{M \times 3 \times 256 \times 256} \rightarrow \mathbb{R}^{m \times d_{256}}$ is an encoder applied instance-level to $[256 \times 256]$ tokens, $\rho: \mathbb{R}^{m \times d} \rightarrow \mathbb{R}^{1 \times d}$ is an aggregation function that pools the embeddings in the bag, and $\zeta: \mathbb{R}^{1 \times d} \rightarrow \mathbb{R}^{\text{\# class}}$ is a bag-level classifier that further processes the bag-level features for downstream slide-level tasks.
\fi

\subsection{Hierarchical Image Pyramid Transformer (HIPT) Architecture}
In adapting ViTs for slide-level representation learning, we reiterate two important challenges distinct from computer vision in natural images: 1) the fixed scale of visual tokens and their hierarchical relationships across image resolutions, and 2) the large sequence lengths of unrolled WSIs. As mentioned, visual tokens in histopathology are generally object-centric (and vary in granularity) across image resolutions, and also have important contextual dependencies such as tumor-immune (inferring favorable prognosis) or tumor-stroma interactions (inferring invasion). Patching with small visual tokens at high objectives ($\textbf{x}_{256}$ at $20\times$) results in large sequence lengths that make self-attention intractable, whereas patching with large visual tokens at low objectives results in loss-of-detail of fine-grained morphological structures ($\textbf{x}_{256}$ at $5\times$) that still requires $[256 \times 256]$ patching at $20\times$. 

To capture this hierarchical structure and the important dependencies that may exist at each image resolution, we approach WSIs similar to long documents as a nested aggregation of visual tokens that recursively break down into smaller tokens until the cell-level (Figure \ref{fig:overview}), written as:
\begin{align*}
    & \operatorname{HIPT}(\mathbf{x}_{\text{WSI}}) = \operatorname{ViT_{WSI}\textrm{-}4096}\big(\big\{ \operatorname{CLS}_{4096}^{(k)} \big\}_{k=1}^{M}\big) \\
    &\rightarrow \operatorname{CLS}_{4096}^{(k)} = \operatorname{ViT_{4096}\textrm{-}256} \big( \{ \operatorname{CLS}_{256}^{(j)} \}_{i=1}^{256} \big) \\
    &\rightarrow \operatorname{CLS}_{256}^{(j)} = \operatorname{ViT_{256}\textrm{-}16} \big( \{ \textbf{x}_{16}^{(i)} \}_{i=1}^{256} \big)
\end{align*}

% $$\textbf{x}_{\text{WSI}} = \Bigg\{ \bigg( \bigg\{ \Big( \{ \textbf{x}_{16}^{(i)} \}_{i=1}^{L} \Big)_{256}^{(j) } \bigg\}_{j=1}^{L} \bigg)_{4096}^{(k)} \Bigg\}_{k=1}^{M}$$

\noindent where 256 is the sequence length of $[16 \times 16]$- and $[256 \times 256]$-patching in $\textbf{x}_{256}$ and $\textbf{x}_{4096}$ images respectively, and $M$ is the total number of $\textbf{x}_{4096}$ images in $\textbf{x}_{\text{WSI}}$. For ease of notation, we refer to $\textbf{x}_{16}$ images as being at the cell-level, $\textbf{x}_{256}$ images as being at the patch-level\footnote{``Patch" is most often used to describe $256 \times 256$ images in pathology, though we note ``patching" an image into smaller images can refer to any resolution.}, $\textbf{x}_{4096}$ images as being at the region-level, with the overall WSI being the slide-level. In choosing these image sizes, the input sequence length of tokens is always $M=256$ in the forward passes for the $\operatorname{ViT_{256}\textrm{-}16}$ and $\operatorname{ViT_{4096}\textrm{-}256}$ (cell- and patch-level aggregation), and usually $M < 256$ in the forward pass for the $\operatorname{ViT_\text{WSI}\textrm{-}4096}$ (slide-level aggregation). The [CLS] tokens from $\operatorname{ViT_{256}\textrm{-}16}$ (the output of the model) are used as the input sequence for $\operatorname{ViT_{4096}\textrm{-}256}$, with the [CLS] tokens from $\operatorname{ViT_{4096}\textrm{-}256}$ subsequently used as the input sequence for $\operatorname{ViT_\text{WSI}\textrm{-}4096}$, with the number of total visual tokens at each stage decreasing geometrically by a factor of 256. In choosing small ViT backbones for each stage, HIPT has less than 10M parameters and is easy-to-implement and train on commercial workstations. We describe each stage below.

% Aggregation of visual tokens is performed across three different stages using ViT backbones, with the number of tokens decreasing geometrically by a factor of $256$ at each stage. 

% %

% where $L = 256$ for $[16 \times 16]$- and $[256 \times 256]$-patching in $\textbf{x}_{256}$ and $\textbf{x}_{4096}$ images respectively, and $M$ is the total number of $\textbf{x}_{4096}$ images in $\textbf{x}_{\text{WSI}}$.

\paragraph{ViT$_{\textbf{256}}$-16 for Cell-Level Aggregation} The computation of $\mathbf{x}_{16}$ cell-level token aggregation within $\mathbf{x}_{256}$ windows follows implementing the vanilla ViT in natural images\cite{dosovitskiy2021image}. Given a $\mathbf{x}_{256}$ patch, the ViT unrolls this image as a sequence of non-overlapping $[16 \times 16]$ tokens followed by a linear embedding layer with added position embeddings to produce a set of 384-dim embeddings $\{\textbf{x}_{16}^{(i)}\}_{i=1}^{256} \in \mathbb{R}^{256 \times 384}$, with a learnable [CLS] token added to aggregate cell embeddings across the sequence. We choose $l = 16$ in this setting to not only follow conventional ViT architectures, but also model important inductive biases in histopathology as at this resolution, a  $[16 \times 16]$ bounding box at $20\times \approx 8 \mu m^2$ area encodes visual concepts that are object-centric in featurizing single cells (\textit{e.g.} - cell identity, shape, roundness).

\paragraph{ViT$_{\textbf{4096}}$-256 for Patch-Level Aggregation} To represent $\mathbf{x}_{4096}$ regions, despite the image resolution being much larger than conventional natural images, the number of tokens remains the same since the patch size scales with the image resolution. From the previous stage, we use $\operatorname{ViT_{256}\textrm{-}16}$ to tokenize non-overlapping $\mathbf{x}_{256}$ patches within each $\mathbf{x}_{4096}$ region, forming the sequence $\{\operatorname{[CLS]}_{256}^{(j)}\}_{j=1}^{256}$ that can be plugged into a ViT block to model larger image contexts. We use a $\operatorname{ViT_{4096}\textrm{-}256(n=4, h=3, d=192)}$ with output $\operatorname{[CLS]}_{4096}$.

\paragraph{ViT$_{\textbf{WSI}}$-4096 for Region-Level Aggregation} In computing the slide-level representation for $\textbf{x}_{\text{WSI}}$, we use a $\operatorname{ViT_{\text{WSI}}\textrm{-}4096(n=2, h=3, d=192)}$ in aggregating the $\operatorname{[CLS]}_{4096}$ tokens. $M$ ranges from $1 - 256$ in our observations depending on size of the WSI. Due to potential tissue segmentation irregularities in patching at $[4096 \times 4096]$, we ignore positional embeddings at this stage. % Instead of using the $\operatorname{[CLS]}_{\text{WSI}}$ token as the slide-level embedding, we use the global attention pooling function from Ilse \etal which similarly pools over the sequence of token representations~\cite{ilse2018attention}.

% \paragraph{Efficient Inference and Batching} At inference time, we feed in $\textbf{x}_{\text{WSI}} \in \mathbb{R}^{M \times 256 \times 256 \times [3 \times 16 \times 16]}$ with a slide-level batch size $B_{\text{WSI}}=1$ into HIPT, in which the respective dimensions correspond to: length of $\mathbf{x}_{4096}$ regions in $\textbf{x}_{\text{WSI}}$, length of $\mathbf{x}_{256}$ patches in $\mathbf{x}_{4096}$,  length of $\mathbf{x}_{16}$ cells in $\mathbf{x}_{256}$, and the remaining dimensions being the $\mathbf{x}_{16}$ shape. Without pre-extracting any tokens, inference can be performed in first defining a $\operatorname{DataLoader}$ class over a $M \times 3 \times 4096 \times 4096$ view of $\textbf{x}_{\text{WSI}}$ with $B_{4096}=1$ to iterate over single $\mathbf{x}_{4096}$ regions, followed by using $\operatorname{Einsum}$ operations to unroll each region via $\operatorname{Einsum}: \mathbb{R}^{1 \times 3 \times 4096 \times 4096} \rightarrow \mathbb{R}^{256 \times 3 \times 256 \times 256}$. Lastly, another $\operatorname{DataLoader}$ is defined over this view using the first dimension as $B_{256}=256$, which then leads to the bottom-up aggregation strategy starting with the $\operatorname{ViT_{256}\textrm{-}16}$ forward pass.

% Moreover, we note though modeling dependencies between tokens are important in WSIs, interactions between very distal tokens are unlikely.

% In adapting ViTs for slide-level representation learning with the aforementioned challenges of 1) modeling the part-whole hierarchy 

\begin{table*}[htbp]
\centering
\begin{tabular}{l|cc|cc|cc}
\toprule
{} & \multicolumn{2}{c|}{\underline{BRCA Subtyping}} & \multicolumn{2}{c|}{\underline{NSCLC Subtyping}} & \multicolumn{2}{c}{\underline{RCC Subtyping}} \\
Architecture & 25\% Training & 100\% Training & 25\% Training & 100\% Training & 25\% Training & 100\% Training \\
\midrule
MIL~\cite{lu2020data}                       &  0.673 $\pm$ 0.112 &  0.778 $\pm$ 0.091 &  0.857 $\pm$ 0.059 &  0.892 $\pm$ 0.042 &  0.904 $\pm$ 0.055 &  0.959 $\pm$ 0.015 \\
CLAM-SB \cite{lu2020data}                        &  0.796 $\pm$ 0.063 &  0.858 $\pm$ 0.067 &  0.852 $\pm$ 0.034 &  0.928 $\pm$ 0.021 &  0.957 $\pm$ 0.012 &  0.973 $\pm$ 0.017 \\
DeepAttnMISL \cite{yao2020whole}                &  0.685 $\pm$ 0.110 &  0.784 $\pm$ 0.061 &  0.663 $\pm$ 0.077 &  0.778 $\pm$ 0.045 &  0.904 $\pm$ 0.024 &  0.943 $\pm$ 0.016 \\
GCN-MIL \cite{zhao2020predicting}                       &  0.727 $\pm$ 0.076 &  0.840 $\pm$ 0.073 &  0.748 $\pm$ 0.050 &  0.831 $\pm$ 0.034 &               0.923 $\pm$ 0.012 &  0.957 $\pm$ 0.012 \\
DS-MIL \cite{li2021dual}                      &  0.760 $\pm$ 0.088 &  0.838 $\pm$ 0.074 &  0.787 $\pm$ 0.073 &  0.920 $\pm$ 0.024 &  0.949 $\pm$ 0.028 &  0.971 $\pm$ 0.016 \\
\rowcolor{maroon!10} HIPT &  \textbf{0.821 $\pm$ 0.069} &  \textbf{0.874 $\pm$ 0.060} &  \textbf{0.923 $\pm$ 0.020} &  \textbf{0.952 $\pm$ 0.021} &  \textbf{0.974 $\pm$ 0.012} &  \textbf{0.980 $\pm$ 0.013} \\
\midrule
$\text{ResNet-50}_{\text{IN}}$ (Mean) &  0.638 $\pm$ 0.089 &  0.667 $\pm$ 0.070 &  0.696 $\pm$ 0.055 &  0.794 $\pm$ 0.035 &  0.862 $\pm$ 0.030 &  0.951 $\pm$ 0.016 \\
$\operatorname{ViT_{256}-16}$ (Mean) &  0.605 $\pm$ 0.092 &  0.725 $\pm$ 0.083 &  0.622 $\pm$ 0.067 &  0.742 $\pm$ 0.045 &  0.848 $\pm$ 0.032 &  0.899 $\pm$ 0.027 \\
\rowcolor{maroon!10} $\operatorname{ViT_{4096}-256}$ (Mean) &  \textbf{0.682 $\pm$ 0.055} &  \textbf{0.775 $\pm$ 0.042} &  \textbf{0.773 $\pm$ 0.048} &  \textbf{0.889 $\pm$ 0.027} &  \textbf{0.916 $\pm$ 0.022} &  \textbf{0.974 $\pm$ 0.016} \\
\bottomrule
\end{tabular}
\caption{\small\textbf{Slide-Level Classification.} \textbf{Top Row.} Ablation study assessing 10-fold cross-validated AUC performance of HIPT across other weakly-supervised architectures. For RCC subtyping, we report the macro-averaged AUC performance across the three subtypes. \textbf{Bottom Row.} Ablation study assessing K-Nearest Neighbors (KNN) performance using the average pre-extracted embeddings.}

\vspace{-4mm}
\label{tab:class}
\end{table*}

\subsection{Hierarchical Pretraining}
\begin{figure}
\begin{center}
\includegraphics[width=.99\linewidth]{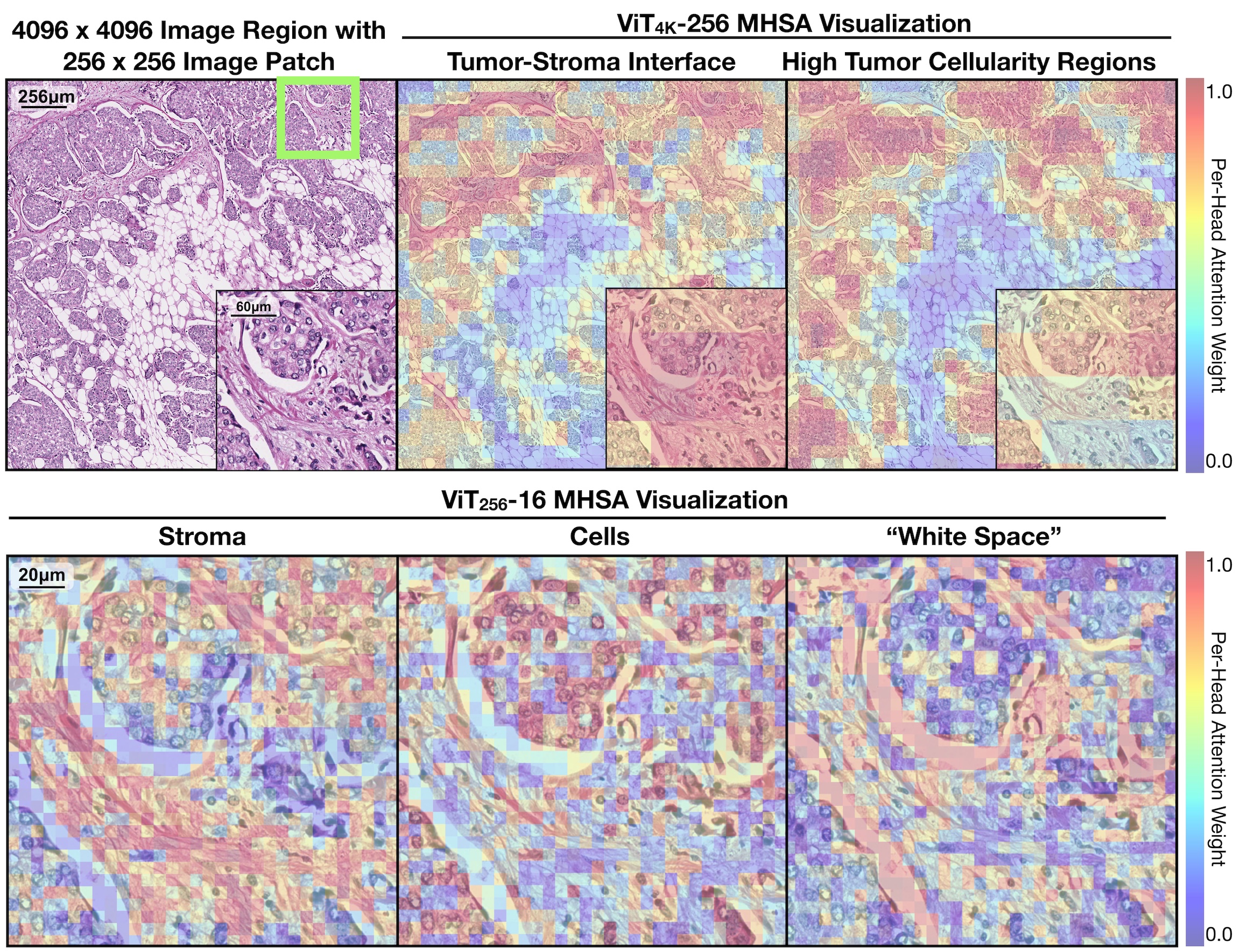}
\end{center}
\vspace{-3mm}
\caption{\small\textbf{Multi-Head Self-Attention Visualization of Self-Supervised ViTs}. For Invasive Ductal Carcinoma (IDC), We show self-supervised visualizations for $\operatorname{ViT_{256}\textrm{-}16}$ and $\operatorname{ViT_{4096}\textrm{-}256}$, pretrained on $\textbf{x}_{256}$ and $\textbf{x}_{4096}$ regions respectively. For $\textbf{x}_{256}$ patches, $\operatorname{ViT_{256}\textrm{-}16}$ is able to delineate stroma, cell, and "white space" presence in $\textbf{x}_{16}$ tokens. For $\textbf{x}_{4096}$ regions, $\operatorname{ViT_{4096}\textrm{-}256}$ delineates coarse-grained morphological features such as tumor nests and their surrounding desmoplastic (loose) stroma.}
\vspace{-5mm}
\label{fig:attention_column}
\end{figure}

In building a MIL framework using only Transformer blocks, we additionally explore and pose a new challenge referred to as slide-level self-supervised learning - which aims at extracting slide-level feature representations in gigapixel images for downstream diagnostic and prognostic tasks. This is an important problem as current slide-level training datasets in CPATH typically have between 100 to 10,000 data points, which may cause MIL methods to overfit due to over-parameterization and lack of labels.\footnote{For rare disease subtypes and clinical trials that study disease progression over the time-course of years, the collection of large patient datasets is difficult to scale for machine learning application.} To address this problem, we hypothesize that the recursive nature of HIPT in using Transformer blocks for image representation learning can enable conventional ViT pretraining techniques (such as DINO \cite{caron2021emerging}) to generalize across stages (of similar subproblems) for high-resolution images. To pretrain HIPT, first, we leverage DINO to pretrain $\operatorname{ViT_{256}\textrm{-}16}$. Then, keeping fixed the weights of $\operatorname{ViT_{256}\textrm{-}16}$, we re-use $\operatorname{ViT_{256}\textrm{-}16}$ as the embedding layer for $\operatorname{ViT_{4096}\textrm{-}256}$ in a second stage of DINO. We refer to this procedure as hierarchical pretraining, which is similarly performed in the context of learning deep belief networks \cite{erhan2010does} and hierarchical transformers for long documents\cite{zhang-etal-2019-hibert}. Though hierarchical pretraining does not reach the slide-level, we show that: 1) pretrained $\textbf{x}_{4096}$ representations in self-supervised evaluation are competitive with supervised methods for slide-level subtyping, and that 2) HIPT with two-stage hiearchical pretraining can reach state-of-the-art performance.

\paragraph{Stage 1: 256 $\times$ 256 Patch-Level Pretraining} To pretrain $\operatorname{ViT_{256}\textrm{-}16}$, we use the the DINO framework for pretraining of $\mathbf{x}_{256}$ patches, in which a student network $\phi_{s_{256}}$ is trained to match the probability distribution of a siamese teacher network $\phi_{t_{256}}$ using a cross-entropy loss $-p_{t_{256}}(\cdot) \operatorname{log} p_{s_{256}}(\cdot)$ with momentum encoding, with $p_{t_{256}}, p_{s_{256}}$ denoting the outputs of $\phi_{t_{256}}(\cdot), \phi_{s_{256}}(\cdot)$ respectively for $\mathbf{\mathbf{x}}_{256}$. As data augmentation for each $\mathbf{x}_{256}$ patch, DINO constructs a set of $M_l = 8$ local views ($\mathbf{x}_{96}$ crops, passed through $\phi_{s_{256}}$) and $M_g = 2$ global views ($\mathbf{x}_{224}$ crops, passed through $\phi_{t_{256}}$) to encourage local-to-global correspondences between the student and teacher, minimizing the function:
$$
\min _{\theta_{s_{256}}} \sum_{\left\{\mathbf{x}_{224}^{(i)}\right\}_{i=1}^{M_g}}^{M_g = 2} \sum_{\left\{\mathbf{x}_{96}^{(j)}\right\}_{j=1}^{M_l}}^{M_l = 8} H \Big( p_{t_{256}}(\mathbf{x}_{224}^{(i)}), p_{s_{256}}\left(\mathbf{x}_{96}^{(j)}\right) \Big)
$$

An intriguing property that makes this data augmentation suitable for histology data is again the natural part-whole hierarchy of cells in a tissue patch. In comparison to natural images in which $[96 \times 96]$ crops may capture only colors and textures without any semantic information, at $20\times$, local $[96 \times 96]$ crops would capture the context of multiple cells and their surrounding extracellular matrices, which has shared mutual information with the broader cellular communities. Similar to the original DINO implementation, we use horizontal flips and color jittering for all views, with solarizing performed on one of the global views.

\paragraph{Stage 2: 4096 $\times$ 4096 Region-Level Pretraining} With the sequence lengths and computational complexity in tokenizing $\mathbf{x}_{4096}$ regions similar to that of $\mathbf{x}_{256}$ patches, we can also borrow an almost identical DINO recipe in also pretraining $\operatorname{ViT_{4096}\textrm{-}256}$ and defining student-teacher networks $\phi_{s_{4096}}(\cdot), \phi_{t_{4096}}(\cdot)$ at this stage. Following extracting $\operatorname{[CLS]}_{256}$ tokens from $\operatorname{ViT_{256}\textrm{-}16}$ as input for $\operatorname{ViT_{4096}\textrm{-}256}$ input, we rearrange $\{\operatorname{[CLS]}_{256}^{(j)}\}_{j=1}^{M=256}$  as a $16 \times 16 \times 384$ 2D feature grid for data augmentations, performing $[6 \times 6], [14 \times 14]$ local-global crops in matching the scale of $[96 \times 96], [224 \times 224]$ crops for $256 \times 256$ inputs. As additional data augmentation, We apply standard dropout ($p=0.10$) to all views following work in Gao \etal\cite{gao-etal-2021-simcse}.

%\subsection{Implementation Details.} 

\section{Experiments}
\vspace{-3mm}
\paragraph{Pretraining:} We pretrain $\operatorname{ViT_{256}\textrm{-}16}$ and $\operatorname{ViT_{4096}\textrm{-}256}$ in different stages, using 10,678 FFPE (formalin-fixed, paraffin-embedded) H\&E-stained diagnostic slides from 33 cancer types in the The Genome Cancer Atlas (TCGA), and extracted 408,218 $\textbf{x}_{4096}$ regions at an $20\times$ objective ($M \approx 38$ regions per slide) for pretraining $\operatorname{ViT_{4096}\textrm{-}256}$, with a total of 104M $\textbf{x}_{256}$ patches for pretraining $\operatorname{ViT_{256}\textrm{-}16}$~\cite{liu2018integrated}. For $\operatorname{ViT_{256}\textrm{-}16}$, we trained for 400,000 iterations using the AdamW optimizer with a batch size of $256$, base learning rate of 0.0005, with the first 10 epochs used to warm up to the base learning rate followed by decay using a cosine schedule. A similar implementation was used for $\operatorname{ViT_{4096}\textrm{-}256}$, with the model trained for 200,000 iterations using the pre-extracted $\operatorname{[CLS}]$ tokens from $\operatorname{ViT_{256}\textrm{-}16}$. %To curate these regions, we used the Tissue Image Analysis (TIA) Toolbox to patch slides into non-overlapping, tissue-containing regions, with additional quality control performed to limit regions that contained predominantly background slide information (\textit{e.g.} - white space). For $\operatorname{ViT_{256}\textrm{-}16}$, we trained for 100 epochs using the AdamW optimizer with a batch size of $256$, base learning rate of 0.0005, with the first 10 epochs used to warm up to the base learning rate followed by decay using a cosine schedule. Following Caron \etal, we also use 0.1 as the temperature for the student, while a linear warmup from 0.04 to 0.07 was used for the temperature for the teacher~\cite{caron2021emerging}. A similar implementation was used for $$\operatorname{ViT_{4096}\textrm{-}256$$, with the model trained for 50 epochs using the pre-extracted $\operatorname{[CLS}$ tokens from $\operatorname{ViT_{256}\textrm{-}16}$. In total, we used approximately $\sim 8 \operatorname{Tb}$ of raw images, which we plan to release to the public using the WebDataset API.

\paragraph{Fine-tuning:} Following hierarchical pretraining, we use the pretrained weights to initialize (and freeze) the $\operatorname{ViT_{256}\textrm{-}16}$ and $\operatorname{ViT_{4096}\textrm{-}256}$ subnetworks, with only a lightweight $\operatorname{ViT_\text{WSI}\textrm{-}4096}$ finetuned. Our work can be viewed as a formulation of MIL that pretrains not only the $[256 \times 256]$ instance-level feature extraction step, but also the downstream aggregation layers which extract coarse-grained morphological features. We finetuned HIPT (and its comparisons) for 20 epochs using the Adam optimizer, batch size of $1$ with $32$ gradient accumulation steps, and a learning rate of $0.01$. For survival prediction, we used the survival cross-entropy loss by Zadeh \& Schmidt\cite{zadeh2020bias}.

\paragraph{Tasks \& Comparisons:} We experiment on several slide-level classification and survival outcome prediction tasks across different organ types in the TCGA~\cite{liu2018integrated}. In comparisons with state-of-the-art weakly-supervised architectures, we tested Attention-Based MIL (ABMIL), and it's variants that use clustering losses (CLAM-SB), clustering prototypes (DeepAttnMISL), modified scoring \& pooling functions (DS-MIL), and graph message passing (GCN-MIL), which used the same hyperparameters as HIPT. Since these methods are agnostic of input features, all comparisons used the pretrained $\operatorname{ViT_{256}\textrm{-}16}$ as instance-level feature extraction. In addition, we also compared variations of HIPT without pretraining and self-attention. Finally, we qualitatively study the attention maps that hierarchical self-supervised ViTs learn in computational histopathology.

% We note that in comparison to survival prediction, subtyping tasks generally require discerning only "instance-level" information as formulated in previous MIL literature, with self-attention in modeling larger context regions not necessary for cancer diagnosis (\textit{e.g} - RCC having distinct subtypes). Still, 
\begin{table*}[h]
\centering
\begin{tabular}{l|cccccc}
\toprule
Architecture &        IDC &    CRC &        CCRCC &        PRCC &        LUAD &        STAD \\
\midrule
ABMIL~\cite{ilse2018attention}   &  0.487 $\pm$ 0.079 &  0.566 $\pm$ 0.075 &  0.561 $\pm$ 0.074 &  \textbf{0.671 $\pm$ 0.076} &  0.584 $\pm$ 0.054 &  0.562 $\pm$ 0.049 \\
DeepAttnMISL~\cite{yao2020whole} &  0.472 $\pm$ 0.023 &  0.561 $\pm$ 0.088 &  0.521 $\pm$ 0.084 &  0.472 $\pm$ 0.162 &  0.563 $\pm$ 0.037 &  0.563 $\pm$ 0.067 \\
GCN-MIL~\cite{li2018graph, zhao2020predicting}   &  0.534 $\pm$ 0.060 &  0.538 $\pm$ 0.049 &  0.591 $\pm$ 0.093 &  0.636 $\pm$ 0.066 &  \textbf{0.592 $\pm$ 0.070} &  0.513 $\pm$ 0.069 \\
DS-MIL~\cite{li2021dual}  &  0.472 $\pm$ 0.020 &  0.470 $\pm$ 0.053 &  0.548 $\pm$ 0.057 &  0.654 $\pm$ 0.134 &  0.537 $\pm$ 0.061 &  0.546 $\pm$ 0.047 \\
\rowcolor{maroon!10} \rowcolor{maroon!10} HIPT   &  \textbf{0.634 $\pm$ 0.050} &  \textbf{0.608 $\pm$ 0.088} &  \textbf{0.642 $\pm$ 0.028} &  0.670 $\pm$ 0.065 &  0.538 $\pm$ 0.044 &  \textbf{0.570 $\pm$ 0.081} \\
\bottomrule
\end{tabular}
\caption{\small\textbf{Survival Prediction.} Ablation study assessing cross-validated c-Index of HIPT across other weakly-supervised architectures.}
\vspace{-4mm}
\label{tab:surv}
\end{table*}

\subsection{Slide-Level Classification}
\vspace{-3mm}
\paragraph{Dataset Description} We follow the study design in ~\cite{lu2020data}; we examined the following tasks evaluated using a 10-fold cross-validated AUC: 1) Invasive Ductal (IDC) versus Invasive Lobular Carcinoma (ILC) in Invasive Breast Carcinoma (BRCA) subtyping, 2) Lung Adenocarcinoma (LUAD) versus Lung Squamous Cell Carcinoma (LUSC) in Non-Small Cell Lung Carcinoma (NSCLC) subtyping, and 3) Clear Cell, Papillary, and Chromophobe Renal Cell Carcinoma (CCRCC vs. PRCC vs. CHRCC) subtyping, with all methods finetuned (for 20 epochs) with varying percentange folds of training data (100\% / 25\%) as data efficiency experiments. Despite RCC subtyping being a relative easy slide-level task due to having distinct subtypes, we ultimately include this task as a benchmark for self-supervised comparisons. % Lastly, we report the performance of RCC subtyping as the macro-averaged AUC across the three classes.

% \paragraph{Baselines} We compare against several set-based network architectures such as Attention-Based MIL (ABMIL), and its variants that use clustering losses (CLAM-SB),  cluster prototypes (DeepAttnMISL), modified scoring \& pooling functions (DS-MIL), and graph message passing as permutation-equivariant aggregation layers (GCN-MIL). % In particular, HIPT shares strong similarity with DS-MIL and DeepGraphConv in learning dependencies of $\mathbf{x}_{256}$ instances in large context regions / overall WSI, however, our approach is unique in that the objective scale across different aggregation layers is preserved. Since these methods are agnostic of input features, we compare all methods using a pretrained $\operatorname{ViT_{256}\textrm{-}16}$ as instance-level feature extraction.

\paragraph{Weakly-Supervised Comparison} Classification results are summarized in Table \ref{tab:class}. Overall, across all tasks and different percentage folds, HIPT consistently achieves the highest macro-averaged AUC performance across all tasks. In comparison with the best performing baseline, CLAM-SB, HIPT achieves a performance increase of 1.86\%, 2.59\%, 0.72\% on BRCA, NSCLC and RCC subtyping respectively using 100\% of training data, with the margin in performance increase widening to 3.14\%, 8.33\%, 1.78\% respectively using 25\% of training data. Similar performance increases are demonstrated on other tasks. HIPT demonstrates the most robust performance when limiting training data, with AUC decreasing slightly from $0.980$ to $0.974$.

\paragraph{K-Nearest Neighbor (KNN)} We take the mean embedding of the pre-extracted embeddings, followed by a KNN evaluation for the above tasks. As a baseline, we use a ResNet-50 pretrained on ImageNet to extract patch-level embeddings. We compare with pre-extracted $\operatorname{ViT_{256}\textrm{-}16}$ patch embeddings from DINO pretraining, and pre-extracted $\operatorname{ViT_{4096}\textrm{-}256}$ region-level embeddings from hierarchical pretraining, with results summarized also in Table \ref{tab:class}. In using the average embedding of each WSI as the ``slide-level representation", we find that $\operatorname{ViT_{4096}\textrm{-}256}$ region-level embeddings in HIPT outperform patch-level embeddings across all tasks, which can be attributed to the broader image contexts used in the WSI for pretraining, and can be intuitively viewed as a closer proxy to the slide-level view than small patches. $\operatorname{ViT_{4096}\textrm{-}256}$ region-level embeddings surpass the AUC performance of weakly-supervised approaches in BRCA and RCC subtyping using 100\% of training data. % In patch-level comparisons, weakly-supervised architectures still outperform ResNet-50 and $\operatorname{ViT_{256}\textrm{-}16}$ embeddings, which can be attributed to $256 \times 256$ patches having smaller image resolutions and thus being able to capture coarser-grained morphological features.

\subsection{Survival Prediction}
\vspace{-3mm}
\paragraph{Dataset Description} For survival outcome prediction, we validated on the IDC, CCRCC, PRCC, and LUAD cancer types which have relatively large sample sizes in the TCGA, in addition to Colon \& Rectal (CRC) and Stomach Adenocarcinoma (STAD) which have been frequently evaluated in real-world clinical studies due to their substantial human intra-observer variability~\cite{echle2020bjc, wulczyn2021npj, skrede2020deep}. All tasks were evaluated using cross-validated concordance index (c-Index).

\paragraph{Weakly-Supervised Comparison} For the following survival prediction tasks in which learning context-aware relationships are important, we observe much larger increases in performance, summarized in Table \ref{tab:surv}. Overall, HIPT achieves the best c-Index performance in the IDC, COADREAD, CCRCC, and STAD cancer types, with the largest improvement demonstrated in IDC (0.634) and COADREAD (0.608) in comparison to other methods. Though other methods such as GCN-MIL use message passing for learning context-aware features, we note that the number of layers needed to achieve similar image receptive fields may cause the number of neighbors to grow exponentially~\cite{li2019deepgcns}. In modeling important long-range dependencies between instances using self-attention across various stages of the hierarchy, the Transformer attention in HIPT is able to capture regional perturbations that have been well characterized as portending worse outcome across different cancer types, as further visualized in Figure ~\ref{fig:attention_column}, \ref{fig:hierarchical_attention}~\cite{abduljabbar2020geospatial, wulczyn2020deep, wulczyn2021npj, skrede2020deep}. % For instance, the so-called epithelial-mesenchymal transition, or tumor-budding, is correlated with higher grade, metastatic potential and worse survival. Other coarse-grained features, such as tumor-immune co-localization or the extent of tumor invasion in healthy stroma, would be difficult to capture using only instance-level feature extraction on $\mathbf{x}_{256}$ images~\cite{saltz2018spatial}. In tissue where normal structures are more likely to be perturbed such as IDC and CRC, HIPT could provide a novel lens for quantifying extent of oncogenic transformation.

% As noted in previous MIL literature, survival prediction is a context-aware task in which long-range dependencies between fine-grained cellular and tissue phenotypes are important in modeling coarse-grained morphological features, such as tumor-immune colocalization or the extend of tumor invasion in healthy stroma.

% Though survival prediction is a more challenging task than cancer subtyping due to the potential for prognostic information existing across the image hierarchy, the trends of performance for HiPT across subtypes show the potential to 
% These performance increases corroborate survival prediction is a more challenging task than cancer subtyping, in which prognostic information may exist across different levels of hierarchy. In particular, we note that amongst 

\subsection{Self-Supervised ViTs Find Unique Morphological Phenotypes}

\begin{figure*}
\begin{center}
\includegraphics[width=1\linewidth]{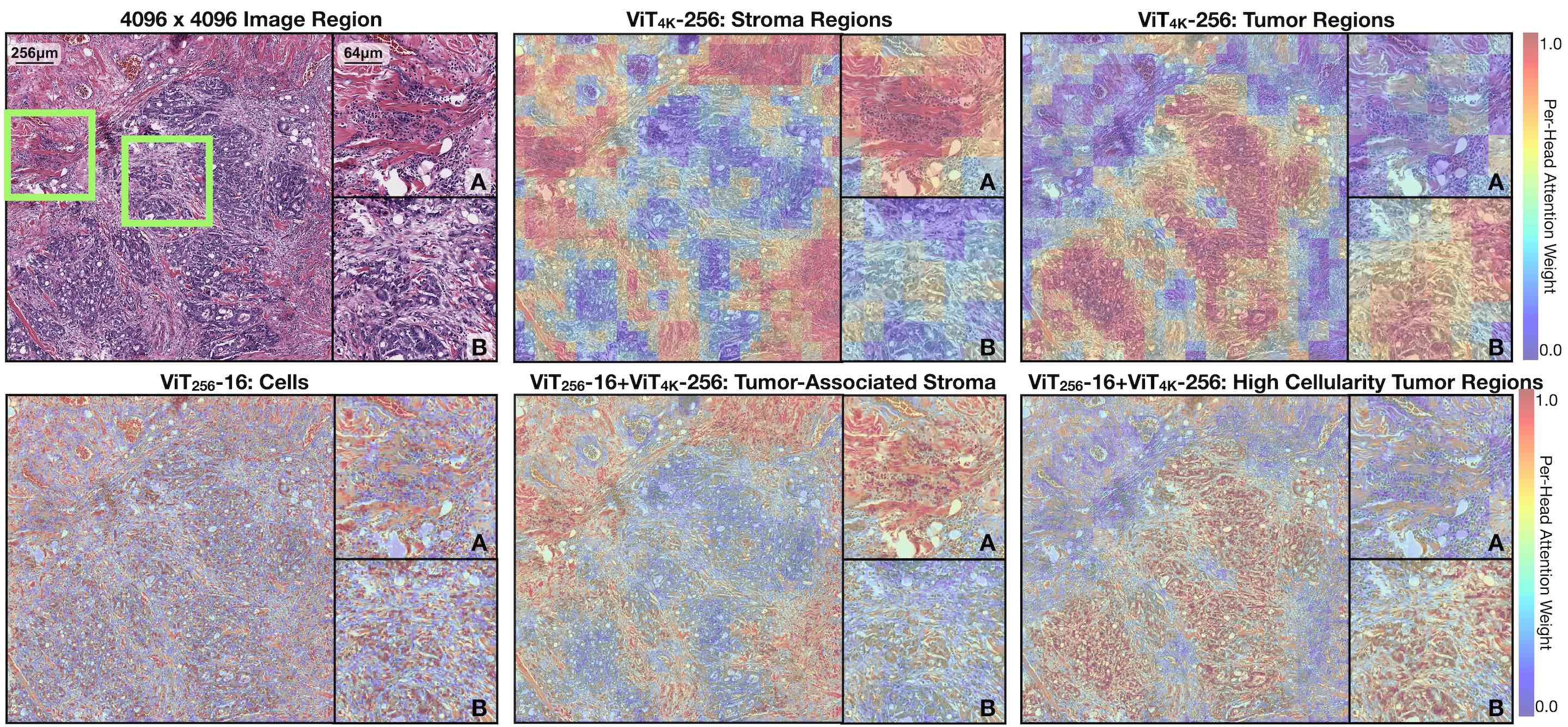}
\end{center}
\vspace{-4mm}
\caption{\textbf{Hierarchical Attention Maps in HIPT}. For Colorectal Cancer (CRC), we observe similar delineation of stroma, cells, and "white space" presence in $\operatorname{ViT_{256}\textrm{-}16}$, and localizing tumor invasion into stroma and muscle (\textbf{A}) and poorly-differentiated glands (\textbf{B}) from $\operatorname{ViT_{4096}\textrm{-}256}$. In factorizing these attention distributions together, we develop hierarchical attention visualizations which can visualize tumor cells with associated stromal tissue and high tumor cellularity regions containing poorly-differentiated glands.}
\vspace{-3mm}
\label{fig:hierarchical_attention}
\end{figure*}

\noindent\textbf{ViT$_{\textbf{256}}$-16 Attention Maps.} For $\mathbf{x}_{256}$ patches, we visualize the different attention heads in MHSA and reveal that ViTs in pathology are able to isolate distinct morphological features. From visual assessment by a board-certified pathologist across several different cancer types, we observe that MHSA in $\operatorname{ViT_{256}\textrm{-}16(n=8, h=6, d=384)}$ captures three distinct fine-grained morphological phenotypes as illustrated in Figure \ref{fig:attention_column}, with general stroma tissue and red blood cells attended in $h=1,2$, cells (normal, atypical, lymphocyte) attended in $h=3,4$, and ``white spaces" (luminal spaces, fat regions, air pockets) attended in $h=5,6$. This observation is in line with current studies that have introspected  self-supervised ViT models, in which the attention heads can be used as a method for object localization or discovery~\cite{caron2021emerging, simeoni2021localizing}. In the application to histopathology tissue, our introspection reveals that the visual tokens at the $[16 \times 16]$ cell-level directly corroborate with semantic, object-centric structures at the $20\times$ objective.

\noindent\textbf{ViT$_{\textbf{4096}}$-256 Attention Maps.} For $\mathbf{x}_{4096}$ regions, we further visualize the attention heads in MHSA from our pretrained $\operatorname{ViT_{4096}\textrm{-}256(n=4, h=6, d=192)}$ model, capturing two distinct coarse-grained phenotypes: tumor-stroma interface attended in $h=1,2,3$, and nested tumor cells and other high tumor cellularity regions in $h=4,5,6$. In comparison with the $\operatorname{ViT_{256}\textrm{-}16}$ attention maps which may capture only nuclear features (\textit{e.g.} - nuclear atypia, shape and size of cells), $\operatorname{ViT_{4096}\textrm{-}256}$ attention maps are able to model the patterns of nested tumor growth, tumor invasion into fat and stroma regions, and other tissue-to-tissue relationships (Figure ~\ref{fig:attention_column}). In factorizing the attention distribution of $[16 \times 16$] cells from $\operatorname{ViT_{256}\textrm{-}16}$ onto highly-attended $[256 \times 256]$ patches from $\operatorname{ViT_{4096}\textrm{-}256}$, we can create a hierarchical attention map, which is able to distinguish tumor cells in stroma tissue from tumor cells in dense tumor cellularity regions (Figure ~\ref{fig:hierarchical_attention}). Overall, these captured coarse- and fine-grained morphological features corroborate with the observed performance increases in both finetuning HIPT in weakly-supervised learning and using averaged HIPT features in KNN evaluation. Additional visualizations are found in the \textbf{Supplement}.
% which corroborate the performance increases from both finetuning HIPT and taking the mean pre-extracted $\operatorname{ViT_{4096}\textrm{-}256}$ embedding in comparison to other methods. Specifically, the 
% the potential in using pre-extracted embeddings as input into self-supervised ViTs that would larger receptive fields or image context.
\subsection{Further Ablation Experiments}
Additional experiments are included in the \textbf{Supplementary Materials}, with main findings highlighted below:\\
\noindent\textbf{The role of pretraining.} Hierarchical pretraining of $\operatorname{ViT_{4096}\textrm{-}256}$ is an important component in our method, as HIPT variants without pretraining overfit in MIL tasks.\\
\noindent\textbf{Comparing patch-level representations.} We assessed quality of other embedding types, and found that $\operatorname{ViT_{256}\textrm{-}16}$ achieves strong representation quality of image patches.\\
\noindent\textbf{Organ-specific versus pan-cancer pretraining}. We additionally assessed the performance of $\operatorname{ViT_{256}\textrm{-}16}$ pretraining on different data distributions, with improved performance in cell localization with pan-cancer pretraining.
\section{Conclusion}
We believe our work is an important step towards self-supervised slide-level representation learning, demonstrating pretrained and finetuned HIPT features achieve superior performance on weakly-supervised and KNN evaluation respectively. Though DINO was used for hierarchical pretraining with conventional ViT blocks, we hope to explore other pretraining methods such as mask patch prediction\cite{dosovitskiy2021image, bao2022beit} and efficient ViT architectures\cite{zhang2021multi, liu2021swin, wang2021pyramid, li2022efficient}.

\noindent\textbf{Limitations:} A limitation of HIPT is the difficulty in pretraining the last aggregation layer due to the small number of WSI data points. In addition, end-to-end hierarchical pretraining of HIPT is computationally intractable on commercial workstations, with pretraining needed to be performed in stages. Lastly, the study design of this work has several constraints, such as: 1) excluded slides in each TCGA cohort due to limited tissue content and difficulty patching at $[4096 \times 4096]$, 2) $\operatorname{ViT_{256}\textrm{-}16}$ pretraining performed on almost all of TCGA and evaluation lacking independent test cohorts, 3), analysis limited to TCGA, which overrepresents patients with European ancestry and not representative of the rich genetic diversity in the world\cite{spratt2016racial}.

\noindent\textbf{Broader Impacts:} Many problems in biology and medicine have hierarchical-like relationships\cite{ma2018using, li2021representation, hinton2021represent}. For instances, DNA motifs within exon sequences which contributes towards protein structure, gene expression, and genetic traits\cite{avsec2021effective, jumper2021highly, demetci2021multi}. Our idea of pretraining neural networks based on hierarchical relationships in large, heterogeneous data modalities to derive a patient- or population-level representation can be extended to other domains.

\section{Acknowledgements}
We thank Felix Yu, Ming Y. Lu, Chunyuan Li, and the BioML group at Microsoft Research New England for their feedback. This work was supported in part by the BWH president's fund, BWH \& MGH Pathology, Google Cloud Research Award, and NIGMS R35GM138216 (F.M.). R.J.C. was also supported by the NSF Graduate Fellowship. T.Y.C. was also supported by the NIH T32CA251062. R.G.K. gratefully acknowledges funding from CIFAR.\\

%Through attention visualizations, we make the finding that pretrained ViT-16, ViT-256 learn important fine- and coarse-grained morphological features that are relevant to context-aware tasks such as survival outcome prediction. 
%Although,, our analysis is limited to the TCGA dataset and 
%In demonstrating success in using unsupervised features for slide-level tasks, our approach has broader implications in proposing objective grading paradigms, as the current standard of care suffers from large intra-observer variabitity within the same stage and grade for many cancer types.

% train in parts
%\fi

% --- supplementary material
\iffalse
\input{sec/X_supplementary}
~
\newpage
~
\newpage
~
\newpage
~
\newpage
~
\newpage
\fi

%%%%%%%%% REFERENCES
{
    %\clearpage
    \small
    \bibliographystyle{ieee_fullname}
    \bibliography{macros,main.bbl}
}

% --- supplementary material
\appendix

% --- PDF will be split by an editor (e.g. macOS preview), so need to restart from page 1
\setcounter{page}{1}

% --- repeat the title (AT: haven't found a more elegant way to do this...)
\twocolumn[
\centering

\Large
\textbf{Supplementary Material} \\
\vspace{1.0em}
] %< twocolumn
\appendix

%\section{Multiple Instance Learning}
%\lorem{2}

\section{Multi-Head Self-Attention}

We briefly introduce self-attention then extend it to the multi-head version. Recall that $ \{\textbf{x}_{l}^{(i)}\}_{i=1}^{M} \in \mathbb{R}^{M \times d_{l}}$ denotes the sequence of visual token extracted from a $L \times L$ patch, where $M$ is the sequence length and $d_{l}$ is the embedding dimension extracted for $l$-sized tokens. For simplicity, we ignore the the subscript $l$ and lower the superscript $i$ to the subscript. The sequence of visual token is re-written as $S = \{\textbf{x}_{i}\}_{i=1}^{M} \in \mathbb{R}^{M \times d_{l}} $

To perform self-attention,  we add three matrices $\boldsymbol{W}_{q}$,$\boldsymbol{W}_{k}$ and $\boldsymbol{W}_{v}$ $\in \mathbb{R}^{d_{l} \times d_{l}}$ used to compute the relationship between the token itself $\textbf{x}_{i}$ and other tokens $\textbf{x}_{j}$ in $S$. Specifically,
\begin{align*}
    \boldsymbol{q}_{i} = \boldsymbol{W}_{q}\textbf{x}_{i},    \boldsymbol{k}_{i} &= \boldsymbol{W}_{k}\textbf{x}_{i},
    \boldsymbol{v}_{i} = \boldsymbol{W}_{v}\textbf{x}_{i}\\
    w_{ij} &= \textrm{softmax}( \frac{\boldsymbol{q}_{i}^{\intercal}\boldsymbol{k}_{j}}{\sqrt{d}_{l}})\\
    \boldsymbol{o}_{i} &= \sum_{j}w_{ij}\boldsymbol{v}_{j},
\end{align*}
where $\boldsymbol{q}_{i}$, $\boldsymbol{k}_{i}$ and $\boldsymbol{v}_{i}$ are query, key and value vector of input token $\textbf{x}_{i}$. Intuitively, each visual token $\textbf{x}_{i}$ computes a similarity score with other tokens and use the normalized similarity score to perform weighted-sum of value vectors of other tokens. The query, key and value vector space are captured by $\boldsymbol{W}_{q}$,$\boldsymbol{W}_{k}$ and $\boldsymbol{W}_{v}$ learned from data. To increase the feature expressiveness, we increase the number of matrices from 1 to $h$ which leads to a set of matrices $\{\boldsymbol{W}^{h^{\prime}}_{q}, \boldsymbol{W}^{h^{\prime}}_{k}, \boldsymbol{W}^{h^{\prime}}_{v} | h^{\prime} = 1,2,3,\dots,h\}$ for the input sequence. The self-attention with $h > 1$ is called multi-head self-attention (MHSA), which we use not only as the building block in permutation-equivariant aggregation of visual tokens, but also demonstrate to learn representations of morphological phenotypes, as seen in Figure~\ref{fig:dino}.

\begin{figure}
\begin{center}
\includegraphics[width=.99\linewidth]{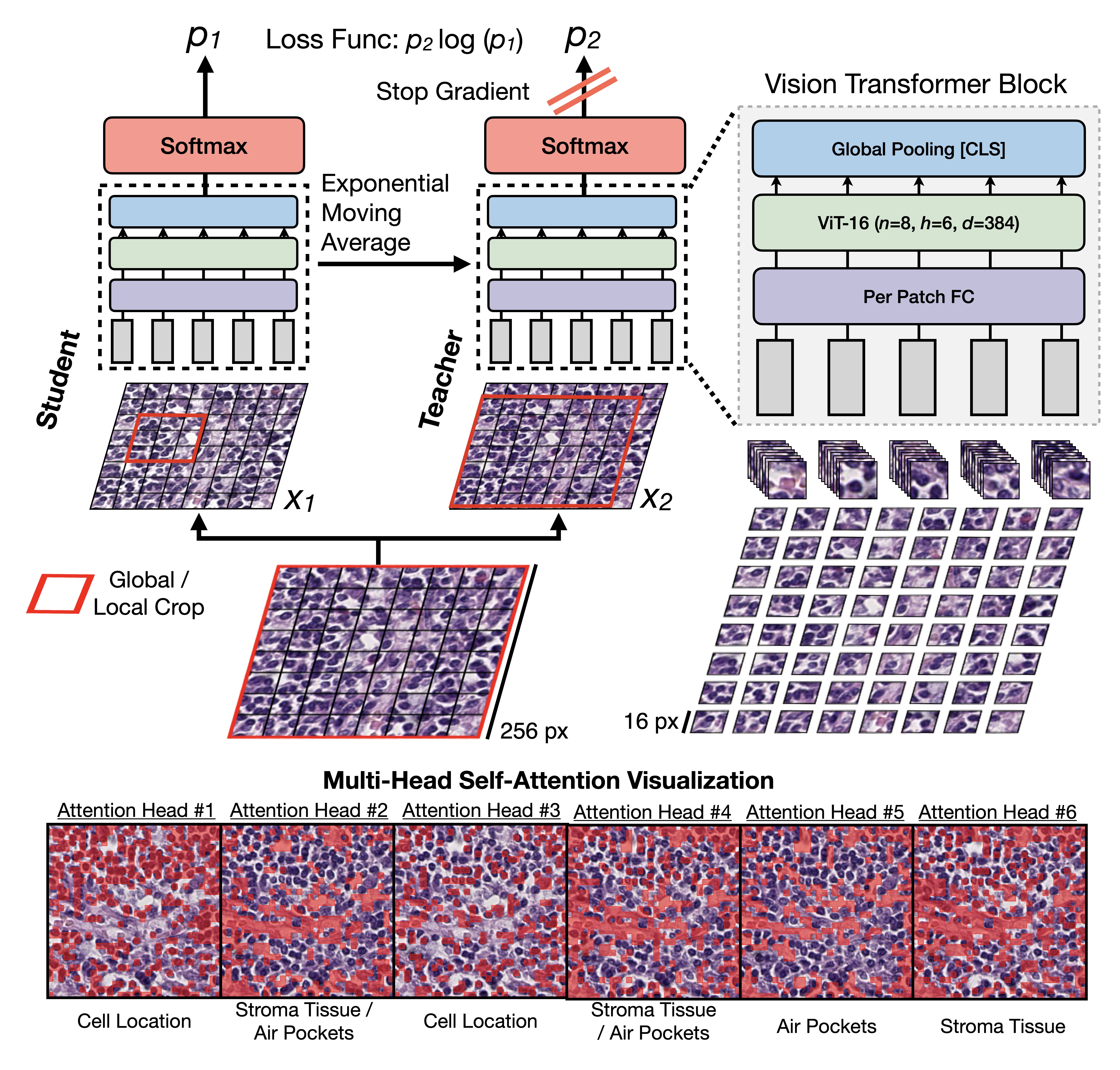}
\end{center}
\vspace{-3mm}
\caption{\small\textbf{$\operatorname{\mathbf{ViT_{256}\textrm{-}16}}$ DINO Pretraining}. Self-supervision using knowledge distillation in DINO to pretrain $\operatorname{ViT_{256}\textrm{-}16}$ on histology image patches~\cite{caron2021emerging}. A student network $\phi_{s_{\text{256}}}$ is trained to match the probability distribution of a Siamese teacher network $\phi_{t_{\text{256}}}$ using a cross-entropy loss, with $\phi$ parameterized using a $\operatorname{ViT_{256}\textrm{-}16}$ model, and local and global crops applied as data augmentation. Interpretability of multi-head attention weights reveal that DINO learns distinct morphological phenotypes. In "red" are high-attention visual tokens with attention weights greater than 0.5.}
\vspace{-5mm}
\label{fig:dino}
\end{figure}

\section{Additional Implementation Details}
\paragraph{Efficient Inference and Batching in HIPT} At inference time, we feed in $\textbf{x}_{\text{WSI}} \in \mathbb{R}^{M \times 256 \times 256 \times [3 \times 16 \times 16]}$ with a slide-level batch size $B_{\text{WSI}}=1$ into HIPT, in which the respective dimensions correspond to: length of $\mathbf{x}_{4096}$ regions in $\textbf{x}_{\text{WSI}}$, length of $\mathbf{x}_{256}$ patches in $\mathbf{x}_{4096}$,  length of $\mathbf{x}_{16}$ cells in $\mathbf{x}_{256}$, and the remaining dimensions being the $\mathbf{x}_{16}$ shape. Without pre-extracting any tokens, inference can be performed in first defining a $\operatorname{DataLoader}$ class over a $M \times 3 \times 4096 \times 4096$ view of $\textbf{x}_{\text{WSI}}$ with $B_{4096}=1$ to iterate over single $\mathbf{x}_{4096}$ regions, followed by using $\operatorname{Einsum}$ operations to unroll each region via $\operatorname{Einsum}: \mathbb{R}^{1 \times 3 \times 4096 \times 4096} \rightarrow \mathbb{R}^{256 \times 3 \times 256 \times 256}$. Lastly, another $\operatorname{DataLoader}$ is defined over this view using the first dimension as $B_{256}=256$, which then leads to the bottom-up aggregation strategy starting with the $\operatorname{ViT_{256}\textrm{-}16}$ forward pass.

\paragraph{Pretraining Dataset Curation:} As noted in the main paper, we pretrain $\operatorname{ViT_{256}\textrm{-}16}$ and $\operatorname{ViT_{4096}\textrm{-}256}$ in different stages using 10,687 FFPE (formalin-fixed, paraffin-embedded) H\&E-stained diagnostic slides from 33 cancer types in the The Genome Cancer Atlas (TCGA), and extracted 408,218 $\textbf{x}_{4096}$ regions at an $20\times$ objective ($M \approx 38$ regions per slide) for pretraining $\operatorname{ViT_{4096}\textrm{-}256}$, with a total of 104 Million $\textbf{x}_{256}$ patches for pretraining $\operatorname{ViT_{256}\textrm{-}16}$~\cite{liu2018integrated}. To curate these regions, we used the Tissue Image Analysis (TIA) Toolbox to patch slides into non-overlapping, tissue-containing regions, with additional quality control performed to limit regions that contained predominantly background slide information (\textit{e.g.} - white space). A limitation of this study is that since our method requires $[4096 \times 4096]$ patching, not all slides from the TCGA were used for pretraining and weakly-supervised evaluation. A full table showing number of WSIs and regions used is shown on a per-cancer basis in Table~\ref{tab:tcga}. %As efficient storage, $\textbf{x}_{4096}$ images for each slide were stored using $\operatorname{TAR}$ archives using the WebDataset API, which we plan to make available in a public release.

\paragraph{Attention Visualization:} To create attention heatmaps, we follow the work of Caron \etal using the attention map for each head at the last ViT stage, and linearly interpolate the attention map such that the attention score each token is $[16 \times 16]$ for $\operatorname{ViT_{256}\textrm{-}16}$ and $[256 \times 256]$ for $\operatorname{ViT_{4096}\textrm{-}256}$. No Gaussian blurring or other smoothing operations were performed. To obtain more granular maps for $\operatorname{ViT_{4096}\textrm{-}256}$, we computed attention scores for each patch using an overlapping stride length of $64$. To create hierarchical attention maps, we factorized the attention weight distribution within each weighted $\textbf{x}_{256}$ patch from $\operatorname{ViT_{4096}\textrm{-}256}$ with the attention weight distribution of $\operatorname{ViT_{256}\textrm{-}16}$. These hierarchical attention maps can be interpreted as: \textit{For weighted $\textbf{x}_{256}$ patches localized by $\operatorname{ViT_{4096}\textrm{-}256}$ by head X, what are the important $\textbf{x}_{16}$ localized within that patch by head Y in $\operatorname{ViT_{256}\textrm{-}16}$?} This allows the creation of certain attention maps that localize: 1) $[16\times 16]$ tumor cells within $256 \times 256$ stromal regions, or 2) $[16\times 16]$ tumor cells within $256 \times 256$ poorly-differentiated glands / larger tumor structures. With $h=6$ for both $\operatorname{ViT_{256}\textrm{-}16}$ and $\operatorname{ViT_{4096}\textrm{-}256}$, a total of 36 hierarchical attention maps can be created. From pathologist inspection, 2-3 heads in each VIT model localized unique morphological phenotypes.

\begin{table}[htbp]
\centering
\begin{tabular}{lrrr}
\toprule
Dataset &  \# Slides &  \# Regions &   Size (in GB) \\
\midrule
ACC      &          223 &        12254 &  230.9 \\
BLCA     &          454 &        21381 &  389.1 \\
BRCA     &         1038 &        30248 &  521.1 \\
CESC     &          271 &         8846 &  152.8 \\
CHOL     &           38 &         2214 &   42.2 \\
COADREAD &          588 &        14705 &  249.3 \\
ESCA     &          145 &         5820 &  103.9 \\
GBMLGG   &         1541 &        54158 &  932.5 \\
HNSC     &          451 &        17204 &  288.6 \\
KIR(C/P/CH)   &          918 &        38019 &  705.5 \\
LIHC     &          375 &        19358 &  369.1 \\
LUADLUSC     &         1008 &        43487 &  757.3 \\
LYM      &           43 &         1590 &   28.3 \\
MESO     &           81 &         2521 &   43.9 \\
OV       &          107 &         5222 &   95.1 \\
PAAD     &          204 &         7377 &  133.2 \\
PRAD     &          421 &        17171 &  301.9 \\
SARC     &          567 &        26974 &  503.7 \\
SKCM     &          456 &        19415 &  351.7 \\
STAD     &          371 &        14664 &  253.0 \\
TGCT     &          233 &         8389 &  154.8 \\
THCA     &          516 &        26611 &  418.5 \\
UCEC     &          561 &        34494 &  628.0 \\
UVM      &           77 &         1657 &   26.7 \\
\midrule
Total & 10687 & 433779 & 7.7 TB \\
\bottomrule
\end{tabular}
\caption{\small\textbf{TCGA Pan-Cancer Datasheet.} Total number of slides, $4096 \times 4096$ image regions, and their storage size.}
\label{tab:tcga}
\end{table}

\section{Additional Quantitative Experiments}

\begin{table*}[htbp]
\footnotesize
\centering
\begin{tabular}{ll|cc|cc|cc}
\toprule
{} & {} & \multicolumn{2}{c|}{\underline{BRCA Subtyping}} & \multicolumn{2}{c|}{\underline{NSCLC Subtyping}} & \multicolumn{2}{c}{\underline{RCC Subtyping}} \\
Architecture & \# Params & 25\% Training & 100\% Training & 25\% Training & 100\% Training & 25\% Training & 100\% Training \\
\midrule
ViT-16$_{\text{PF}}$, AP-256, AP-4096 & 494597      &  0.784 $\pm$ 0.061 &  0.837 $\pm$ 0.062 &  0.835 $\pm$ 0.050 &  0.928 $\pm$ 0.023 &  0.955 $\pm$ 0.016 &  0.965 $\pm$ 0.013 \\
ViT-16$_{\text{PF}}$, ViT-256, ViT-4096 & 3388996    &  0.758 $\pm$ 0.076 &  0.823 $\pm$ 0.071 &  0.695 $\pm$ 0.069 &  0.786 $\pm$ 0.096 &                0.928 $\pm$ 0.038 &  0.956 $\pm$ 0.016 \\
ViT-16$_{\text{PF}}$, ViT-256$_{\text{P}}$, ViT-4096 & 3388996  &  0.762 $\pm$ 0.089 &  0.827 $\pm$ 0.069 &  0.652 $\pm$ 0.076 &  0.820 $\pm$ 0.047 &  0.935 $\pm$ 0.022 &  0.956 $\pm$ 0.013 \\
ViT-16$_{\text{PF}}$, ViT-256$_{\text{PF}}$ & 505204 &  \textbf{0.821 $\pm$ 0.069} &  \textbf{0.874 $\pm$ 0.060} &  \textbf{0.923 $\pm$ 0.020} &  \textbf{0.952 $\pm$ 0.021} &  \textbf{0.974 $\pm$ 0.012} &  \textbf{0.980 $\pm$ 0.013} \\
\midrule
$\text{ResNet-50}_{\text{B3, IN}}$, GMP & - &  0.638 $\pm$ 0.089 &  0.667 $\pm$ 0.070 &  0.696 $\pm$ 0.055 &  0.794 $\pm$ 0.035 &  0.862 $\pm$ 0.030 &  0.951 $\pm$ 0.016 \\
ViT-16$_{\text{PF}}$, GMP & - &  0.605 $\pm$ 0.092 &  0.725 $\pm$ 0.083 &  0.622 $\pm$ 0.067 &  0.742 $\pm$ 0.045 &  0.848 $\pm$ 0.032 &  0.899 $\pm$ 0.027 \\
ViT-16$_{\text{PF}}$, ViT-256$_{\text{PF}}$, GMP & - &  \textbf{0.682 $\pm$ 0.055} &  \textbf{0.775 $\pm$ 0.042} &  \textbf{0.773 $\pm$ 0.048} &  \textbf{0.889 $\pm$ 0.027} &  \textbf{0.916 $\pm$ 0.022} &  \textbf{0.974 $\pm$ 0.016} \\
\bottomrule
\end{tabular}
\caption{\small\textbf{Slide-Level Classification.} \textbf{Top Row.} Ablation study assessing impact of Transformer attention, pretraining, and parameter freezing in the HIPT architecture. \textbf{Bottom Row.} Ablation study assessing K-Nearest Neighbors (K-NN) performance using the average pre-extracted embeddings with different pretrained embedding types. \textbf{Abbreviations.} "P" = Pretrained. "F" = Frozen. "PF" = Pretrained and Frozen. "ViT" = Vision Transformer. "ResNet-50$_{\text{B3, IN}}$ = ResNet-50 truncated after the 3rd residual block, with ImageNet transfer learning. "AP" = Attention Pooling only. "GMP" = Global Mean Pooling only. For ease of notion and table space, we remove the subscript which denotes the image resolution operated on by the ViT.}
\label{tab:supp_class}
\end{table*}

\iffalse
\begin{table*}[htbp]
\footnotesize
\centering
\begin{tabular}{l|ccccccccc}
\toprule
Method & Adi &  Back &    Deb &    Lym &    Muc &   Str &   Norm & Tum &    All \\
\midrule
$\text{ResNet-50}_{\text{B3, IN}}$     &  0.983 &  1.000 &  0.997 &  0.974 &  0.963 &  0.988 &  0.982 &  0.978 &  0.983 \\
ViT-16$_{\text{PF, BRCA, S1}}$   &  0.999 &  1.000 &  0.999 &  0.985 &  0.992 &  0.960 &  0.992 &  0.967 &  0.987 \\
ViT-16$_{\text{PF, PANC, S1}}$    &  0.997 &  1.000 &  0.995 &  0.981 &  0.982 &  0.966 &  0.980 &  0.960 &  0.983 \\
ViT-16$_{\text{PF, PANC, S4}}$ &  0.997 &  1.000 &  0.994 &  0.986 &  0.983 &  0.971 &  0.985 &  0.963 &  0.985 \\
\bottomrule
\end{tabular}
\caption{Caption}
\label{tab:supp_kather}
\end{table*}
\fi

\begin{table*}[htbp]
\footnotesize
\centering
\begin{tabular}{ll|cccc}
\toprule
Method & Dim & CRC-100K-R $\uparrow$ & CRC-100K-N $\uparrow$ & BCSS $\uparrow$ & BreasthPathQ $\downarrow$ \\
\midrule
$\text{ResNet-50}_{\text{B3, IN}}$ & 1024 & 0.935 & 0.983 & 0.599 & 0.058 \\
ViT-16$_{\text{PF, BRCA, S1}}$ & 384 & 0.941 & \textbf{0.987} & 0.593 & 0.029 \\
ViT-16$_{\text{PF, PANC, S1}}$ & 384 & \textbf{0.941} & 0.983 & \textbf{0.616} & \textbf{0.023} \\
ViT-16$_{\text{PF, PANC, S4}}$ & 1536 & 0.927 & 0.985 & 0.612 & 0.052 \\
\bottomrule
\end{tabular}
\caption{\small\textbf{Patch-Level Classification.} Ablation study assessing K-Nearest Neighbors (K-NN) performance on patch-level datasets with different embedding types. \textbf{Abbreviations.} "P" = Pretrained. "F" = Frozen. "PF" = Pretrained and Frozen. "ViT" = Vision Transformer. "ResNet-50$_{\text{B3, IN}}$ = ResNet-50 truncated after the 3rd residual block, with ImageNet transfer learning. "BRCA" = Pretrained on BRCA only. "PANC" = Pretrained on Pan-Cancer data. "S1" = Using features from only last stage / hidden layer of ViT. "S4" = Featured concatenated from the last 4 stages. For ease of notion and table space, we remove the subscript which denotes the image resolution operated on by the ViT.}
\label{tab:supp_kather}
\end{table*}

\subsection{Variations in HIPT Architecture}
\paragraph{Model Descriptions} We performed ablation experiments assessing the most impactful components of the HIPT architecture, as demonstrated in the slide-level classification results in Table \ref{tab:supp_class}. Specifically, we inspected variations in the HIPT architecture:

\begin{itemize}
    \item $\operatorname{\mathbf{ViT_{256}\textrm{-}16}}_{\textbf{PF}}$, \textbf{AP-256, AP-4096}: This variation uses no ViT components as permutation-equivariant aggregation hidden layers at the patch- and region-level. Features are still pre-extracted using a $\operatorname{ViT_{256}\textrm{-}16}$ (denoted with ``P" for pretrained, ``F" for frozen), however, only attention pooling (AP) is performed across the different stages.
    \item $\operatorname{\mathbf{ViT_{256}\textrm{-}16}_{\textbf{PF}}}$, $\operatorname{\mathbf{ViT_{4096}\textrm{-}256}}$, $\operatorname{\mathbf{ViT_\text{WSI}\textrm{-}4096}}$: This variation uses $\operatorname{ViT_{4096}\textrm{-}256}$ and $\operatorname{ViT_\text{WSI}\textrm{-}4096}$ for patch- and region-level aggregation, however, these aggregation layers are trained from scratch.
    \item $\operatorname{\mathbf{ViT_{256}\textrm{-}16}}_{\textbf{PF}}$, $\operatorname{\mathbf{ViT_{4096}\textrm{-}256}}_{\textbf{P}}$, $\operatorname{\mathbf{ViT_\text{WSI}\textrm{-}4096}}$: This variation uses $\operatorname{ViT_{4096}\textrm{-}256}$ and $\operatorname{ViT_\text{WSI}\textrm{-}4096}$ for patch- and region-level aggregation, with$\operatorname{ViT_{4096}\textrm{-}256}$ additionally pretrained using Stage 2 Hierarchical Pretraining.
     \item $\operatorname{ViT_{256}\textrm{-}16}_{\text{PF}}$, $\operatorname{ViT_{4096}\textrm{-}256}_{\text{PF}}$, $\operatorname{ViT_\text{WSI}\textrm{-}4096}$: This variation uses $\operatorname{ViT_{4096}\textrm{-}256}$ and $\operatorname{ViT_\text{WSI}\textrm{-}4096}$ for patch- and region-level aggregation, with $\operatorname{ViT_{4096}\textrm{-}256}$ additionally pretrained and frozen. Only parameters for $\operatorname{ViT_\text{WSI}\textrm{-}4096}$ are finetuned.
\end{itemize}

\paragraph{Pretraining and Freezing Prevents Overfitting} Across slide-level classification tasks, we observe that pretraining and freezing $\operatorname{ViT_{4096}\textrm{-}256}$ in HIPT is an important component an achieving strong performance. Without freezing, training the parameters for $\operatorname{ViT_{4096}\textrm{-}256}$ results in the total number of trainable parameters to be 3388996. Though small for a Transformer model, training and finetuning on WSI datasets with less than 1000 data points may easily result in overfitting, as demonstrated in Table~\ref{tab:supp_class}. In particular, we observe that without freezing, performance drops from 0.923 to 0.652 and 0.952 to 0.820 on NSCLC subtyping with 25\% and 100\% training data finetuning respectively. We also compare HIPT to a variation that does not use any Transformer attention in patch- and region-level aggregation, which performed better than HIPT variations with $\operatorname{ViT_{4096}\textrm{-}256}$ finetuned, but still worse than HIPT with $\operatorname{ViT_{4096}\textrm{-}256}$ frozen.

\subsection{Assessing Quality of $\textbf{x}_{256}$ Representations} Though the primary focus of our paper is in hierarchical pretraining using the HIPT architecture, we also make publicly available the pretrained weights of $\operatorname{ViT_{256}\textrm{-}16}$, which can be used as a general feature extractor for $256 \times 256$ histology patches. Accordingly, we perform a model audit that assesses the quality of $\textbf{x}_{256}$ representations for our pan-cancer $\operatorname{ViT_{256}\textrm{-}16}$ model (denoted as $\operatorname{ViT_{256}\textrm{-}16}_{\text{PF, PANC, S1}}$), and compare with three other embedding types: 1) Pretrained ImageNet (IN) features from a truncated ResNet-50 (after the 3rd residual block, or $\text{ResNet-50}_{\text{B3, IN}}$), 2) $\operatorname{ViT_{256}\textrm{-}16}_{\text{PF, BRCA, S1}}$ features trained on only data from TCGA-BRCA (as an organ-specific comparison using the same hyper-parameters and \# of iteration), and 3) $\operatorname{ViT_{256}\textrm{-}16}_{\text{PF, PANC, S4}}$ features trained on pan-cancer data from the TCGA (but with features concatenated across the last four hidden layers, denoted as "S4", versus just the last stage, "S1"). We assess these representations quantitatively using KNN evaluation for slide-level tasks (Table~\ref{tab:supp_class}) and most patch-level tasks (using global mean pooling (GMP), Table~\ref{tab:supp_kather}), as well as qualitatively using UMAP scatter-plots. Description of patch-level datasets are found below:

\begin{figure*}[t]
\begin{center}
\includegraphics[width=.99\linewidth]{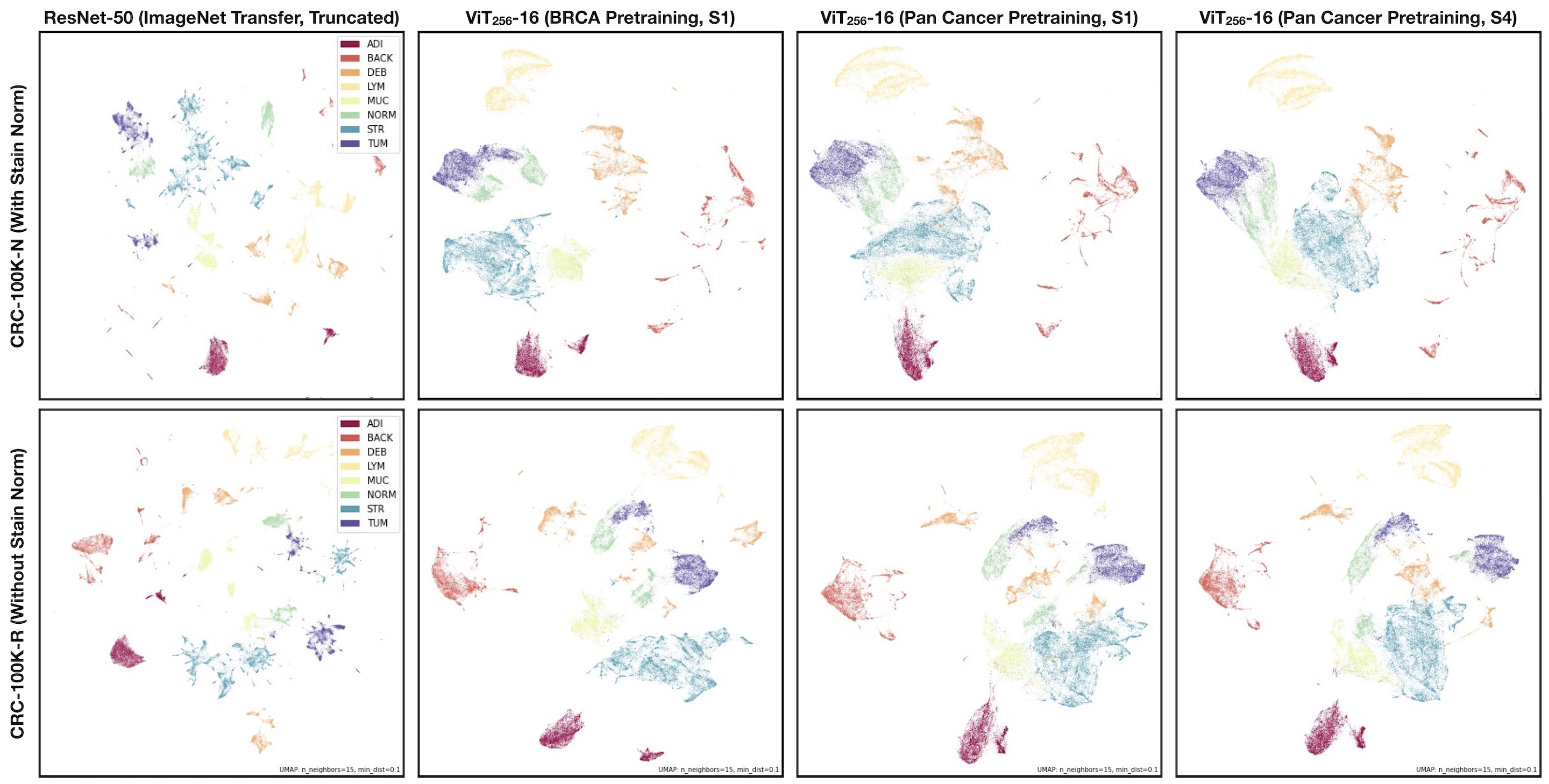}
\end{center}
\vspace{-4mm}
\caption{\textbf{UMAP Visualization of Pretrained Embeddings}. 2D UMAP scatter-plot visualizing global structure of pretrained embeddings on CRC-100K (with and without Macenko stain normalization). In ResNet-50$_{\text{B3, IN}}$, global structure for many class types are not well-preserved, with worse representation quality in CRC-100K without stain normalization. Across all $\operatorname{ViT_{256}\textrm{-}16}$ models, global structures for morphological subtypes such as normal, tumor, stroma, and mucous are well-preserved in both datasets. We used default UMAP parameters of: $\operatorname{neighbors}=15, \operatorname{dist}=0.1$.}
\vspace{-3mm}
\label{fig:umap}
\end{figure*}

\begin{itemize}
    \item \textbf{CRC-100K:} CRC-100K is a dataset of 100,000 histological images of human colorectal cancer and healthy tissue, extracted as $224 \times 224$ patches at $20\times$ magnification (available with and without Macenko normalization), and is annotated with the following non-overlapping tissue classes: adipose (Adi), background (Back), debris (Deb), lymphocytes (Lym), mucus (Muc), smooth muscle (Mus), normal colon mucosa (Norm), cancer-associated stroma (Str), colorectal adenocarcinoma epithelium (Tum)~\cite{kather2016multi}. We experiment on CRC-100K with and without Macenko Normalization (denoted with ``-N" and ``-R" respectively). We report multiclass AUC performance.
    \item \textbf{BCSS}: The Breast Cancer Semantic Segmentation (BCSS) Dataset is a dataset that contains over 20,000 segmentation annotations from the TCGA-BRCA cohort, from which we mined $256 \times 256$ patches at $20\times$ magnification for the following overlapping cell types: background tissue, tumor cells, stroma cells, and lymphocyte infiltrates~\cite{amgad2019structured}. Unlike CRC-100K, BCSS overlaps in label categories as tissue patches can have multiple labels of each cell type. As a result, we used the majority cell phenotype as the patch-level label during supervision. We report multiclass AUC performance.
    \item \textbf{BreastPathQ:} BreastPathQ is a challenge dataset from the TCGA-BRCA cohort that measures tumor cellularity, which measures the fractional occupancy of tumor cell presence in the image patch~\cite{petrick2021spie}. We evaluated on the public train/validation split of the challenge, which provides 2579 and 187 patches respectively at $20\times$, and report mean-squared error (MSE) using linear regression.
\end{itemize}

\paragraph{Comparison with ImageNet Features} In comparing $\operatorname{ViT_{256}\textrm{-}16}_{\text{PF, PANC, S1}}$ with $\text{ResNet-50}_{\text{B3, IN}}$, on both patch- and slide-level evaluation using KNN, we observe that $\operatorname{ViT_{256}\textrm{-}16}_{\text{PF, PANC, S1}}$ features are generally more robust. On patch-level classification, $\text{ResNet-50}_{\text{B3, IN}}$ and $\operatorname{ViT_{256}\textrm{-}16}_{\text{PF, PANC, S1}}$ do equally well on CRC-100K-N (MNIST equivalent of patch-level tasks in CPATH), with $\operatorname{ViT_{256}\textrm{-}16}_{\text{PF, PANC, S1}}$ performing better on BCSS and BreastPathQ (more challenging datasets with noisy and fine-grained labels respectively). $\operatorname{ViT_{256}\textrm{-}16}_{\text{PF, PANC, S1}}$ also performed better on CRC-100K-R, in which images are not stain normalized and thus have more variation due to institution-specific staining protocols. One hypothesis for the surprisingly robust $\text{ResNet-50}_{\text{B3, IN}}$ features is that the feature maps before the last residual block are low-level feature descriptors, and are able to distinguish between distinct morphologies such as tumor versus stroma, or tumor versus adipose tissue. In visualizing UMAP scatter plots of pre-extracted $\text{ResNet-50}_{\text{B3, IN}}$ features, despite the high AUC performance on CRC-100K-R and CRC-100K-N, the representation quality is poor as global structures within the same class types are not preserved (Figure \ref{fig:umap}). On the other hand, global structures for classes such as stroma, tumor, normal, and mucous tissue are well-preserved for all ViT models. 

In slide-level KNN evaluation, interestingly, we see that mean $\text{ResNet-50}_{\text{B3, IN}}$ features out-perform mean $\operatorname{ViT_{256}\textrm{-}16}_{\text{PF, PANC, S1}}$ features on NSCLC and RCC subtyping, with $\operatorname{ViT_{256}\textrm{-}16}_{\text{PF, PANC, S1}}$ performing better on BRCA subtyping. This result can be attributed to NSCLC and RCC subtyping being generally easier tasks in which subtypes can be more readily distinguished, whereas BRCA subtyping being more challenging due to the phenotypic similarity of IDC to ILC that typically requires stroma context.

\begin{figure*}[t]
\begin{center}
\includegraphics[width=.99\linewidth]{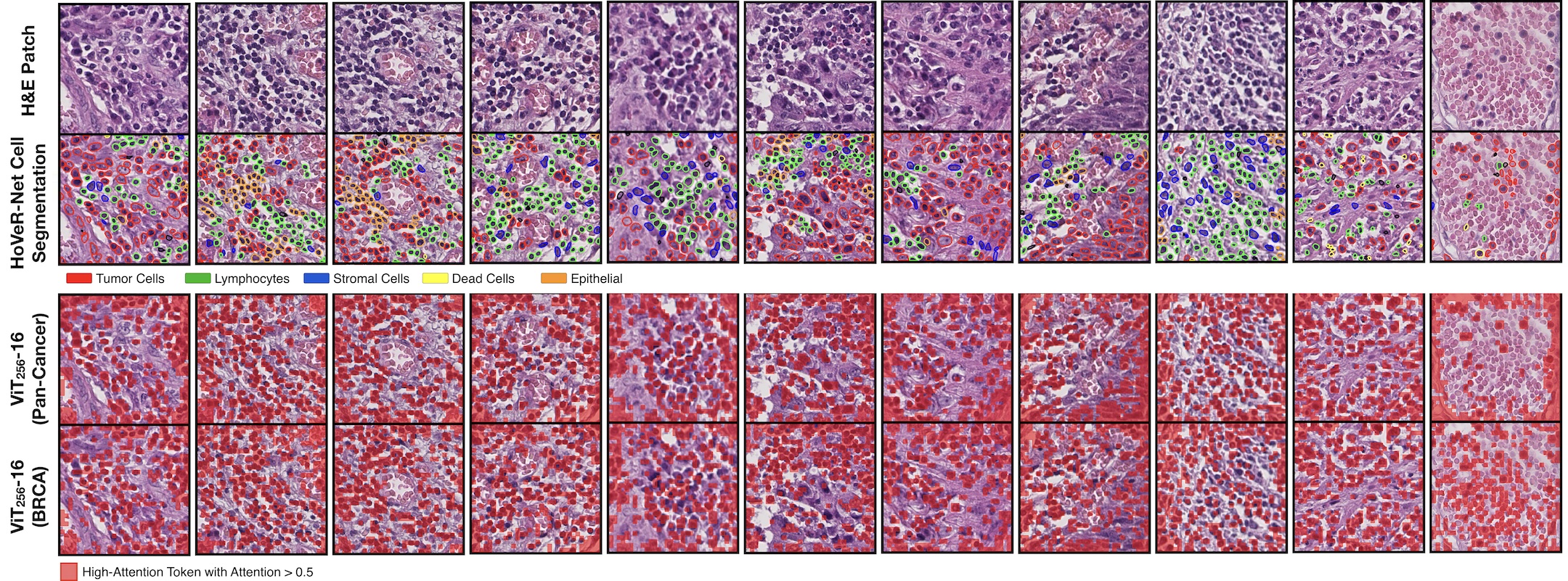}
\end{center}
\vspace{-4mm}
\caption{\textbf{Comparison of Pan-Cancer versus BRCA pretraining in $\operatorname{\mathbf{ViT_{256}\textrm{-}16}}$}. For $\operatorname{ViT_{256}\textrm{-}16}_{\text{PF, BRCA, S1}}$ and ViT-16$_{\text{PF, PANC, S1}}$, we visualize the attention weights for $h=0, 3$ respectively which we observe to be good at localizing cells. For both $\operatorname{ViT_{256}\textrm{-}16}$  models, overlayed in "red" are high-attention visual tokens with attention weights greater than 0.5. In addition, we all show accompanying cell segmentation results from a HoVeR-Net model~\cite{graham2019hover}. Overall, we observe that Pan-Cancer pretraining in $\operatorname{ViT_{256}\textrm{-}16}$ is able to localize tumor and lymphocytes better than BRCA pretraining, with BRCA pretraining attending to blood cells in some instances.}
\vspace{-3mm}
\label{fig:cellseg}
\end{figure*}

\paragraph{Organ-Specific versus Pan-Cancer Image Pretraining} In comparing $\operatorname{ViT_{256}\textrm{-}16}_{\text{PF, PANC, S1}}$ with $\operatorname{ViT_{256}\textrm{-}16}_{\text{PF, BRCA, S1}}$, briefly, we note that both models have similar performance across most patch-level tasks, with $\operatorname{ViT_{256}\textrm{-}16}_{\text{PF, PANC, S1}}$ performing slightly better on BCSS and BreastPathQ evaluation. In examining global structure of morphological subtypes in CRC-100K, despite not being pretrained on CRC data, $\operatorname{ViT_{256}\textrm{-}16}_{\text{PF, BRCA, S1}}$ is able to preserve global structure better in stroma and mucous subtypes. In Figure~\ref{fig:cellseg}, we additionally visualize the cell localization results of each ViT model, with $\operatorname{ViT_{256}\textrm{-}16}_{\text{PF, PANC, S1}}$ performing overall better in localizing tumor cells and lymphocytes, with $\operatorname{ViT_{256}\textrm{-}16}_{\text{PF, BRCA, S1}}$ attending more to blood cells.

\paragraph{Feature Concatenation across $\operatorname{\mathbf{ViT_{256}\textrm{-}16}}$ Hidden Layers} Following Caron \etal, we also concatenated the $\operatorname{[CLS]}$ tokens from the last four stages of $\operatorname{ViT_{256}\textrm{-}16}$, resulting in a 1536-dim embedding~\cite{caron2021emerging}. We observe no improvement in performing feature concatenation.

\paragraph{Concluding Remarks on $\textbf{x}_{256}$ Representations} In addition to having strong representation quality and interpretability mechanisms in finding unique morphological features, a key detail about $\operatorname{ViT_{256}\textrm{-}16}$ is that the embedding dimension is relatively small (with a length of 384), which allows Stage 2 Hierarchical Pretraining and finetuning of $\operatorname{ViT_{4096}\textrm{-}256}$ to be tractable on commercial workstations. Though longer embedding dimensions may lead to better performance on some patch-level tasks, our ultimate goal is in building hierarchical models for slide-level representation learning, which relies on: 1) shorter embedding dimensions, and 2) $[16 \times 16]$ cell-level interpretability.

\section{Additional Visualizations}

Additional visualizations for hierarchical attention maps are shown in Figure~\ref{fig:brca},~\ref{fig:crc}. Attached in is also a hierarchical attention map for Figure 3 in its native $4096 \times 4096$ resolution. Due to space constraints, we created a repository to visualize $\operatorname{ViT_{256}\textrm{-}16}$, $\operatorname{ViT_{4096}\textrm{-}256}$, and hierarchical attention heatmaps at the following link address: \href{https://bit.ly/HIPT-Supplement}{https://bit.ly/HIPT-Supplement}.

%\noindent\textbf{HuggingFace Demo.} To visualize attention maps for your own images, we also provide a link to the following demo made available with the Gradio API on Hugging Face: \href{https://huggingface.co/spaces/Anon4review/HIPTDemo}{https://huggingface.co/spaces/Anon4review/HIPTDemo}.

\begin{figure*}[t]
\begin{center}
\includegraphics[width=.99\linewidth]{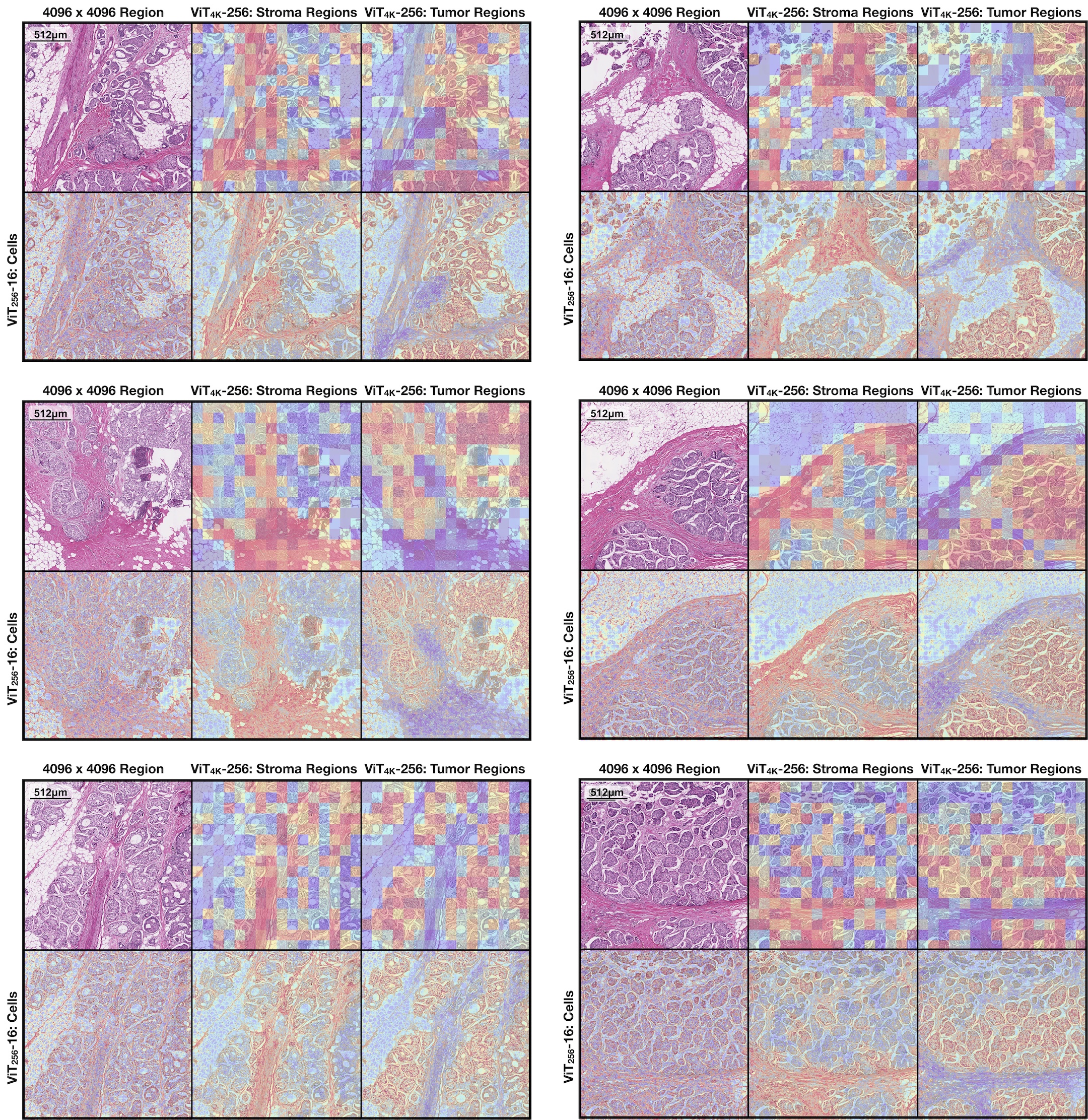}
\end{center}
\vspace{-4mm}
\caption{\textbf{Hierarchical Attention Maps for Invasive Breast Carcinoma (BRCA)}. Similar to Figure~\ref{fig:hierarchical_attention}, factorized attention distributions of combined $\operatorname{ViT_{256}\textrm{-}16}$  and $\operatorname{ViT_{4096}\textrm{-}256}$  attention distributions are able to localize: 1) invasive tumor cells in demoplastic stroma, 2) tumor cells arranged in larger tumor nest patterns, which is important in distinguishing Invasive Ductal versus Lobular Carcinoma as well as survival outcomes.}\vspace{-3mm}
\label{fig:brca}
\end{figure*}

\begin{figure*}[t]
\begin{center}
\includegraphics[width=.99\linewidth]{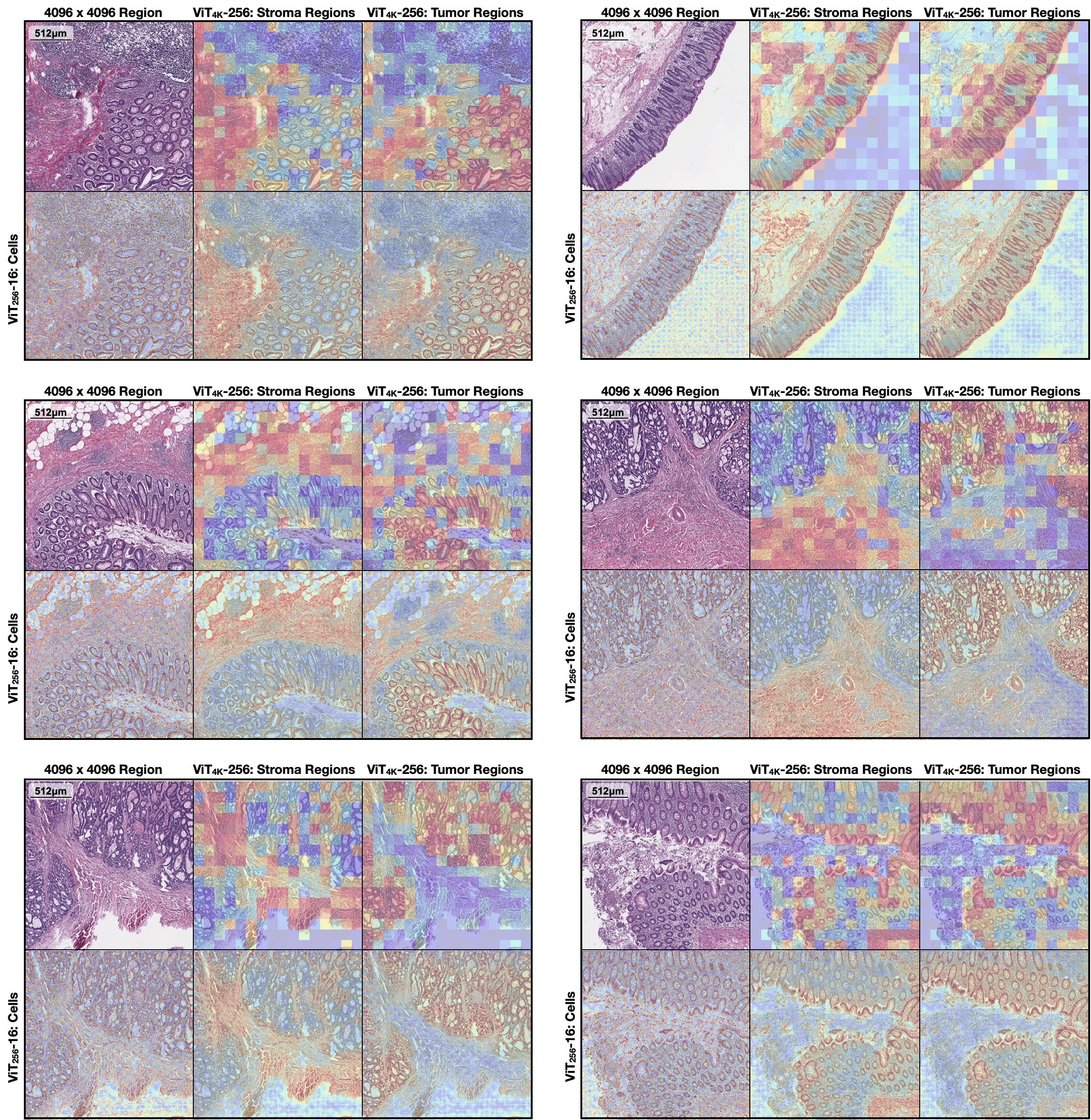}
\end{center}
\vspace{-4mm}
\caption{\textbf{Hierarchical Attention Maps for Colorectal Cancer (CRC)}. Similar to Figure~\ref{fig:hierarchical_attention}, factorized attention distributions of combined $\operatorname{ViT_{256}\textrm{-}16}$ and $\operatorname{ViT_{4096}\textrm{-}256}$ attention distributions are able to localize: 1) invasive tumor cells in muscle and stromal regions, 2) tumor cells forming poorly-differentiated glands, which are both important prognostic histopathologic biomarkers in determining severity in cancer staging and survival outcomes.}\vspace{-3mm}
\label{fig:crc}
\end{figure*}

% --- uncomment this to read the instructions
%\input{sec/X_instructions}

\end{document}